\documentclass[runningheads]{llncs}
\usepackage[T1]{fontenc}
\usepackage[nocompress]{cite}

\usepackage{amsmath,amssymb,amsfonts}
\usepackage{lipsum}

\usepackage{algorithm}
\usepackage{algpseudocode}
\usepackage{graphicx, stfloats}
\usepackage{caption}
\usepackage{subcaption}
\usepackage{multicol}
\usepackage{adjustbox}
\usepackage{float}

\usepackage{hhline}
\usepackage{makecell}
\usepackage{multirow}
\usepackage{tabulary}
\usepackage{float}
\usepackage{adjustbox}

\usepackage{textcomp}
\usepackage{xcolor}

\usepackage[hidelinks]{hyperref}

\begin{document}
\title{Analytics Modelling over Multiple Datasets using Vector Embeddings}
%
%
\author{Andreas Loizou \and Dimitrios Tsoumakos}
\institute{\textit{Database and Knowledge Systems Lab} \\
\textit{School of ECE, National Technical University of Athens}\\
\email{\{antreasloizou,dtsouma\}@mail.ntua.gr}}

%
%
%
\maketitle              
\vspace{-0.25in}
\begin{abstract}
The massive increase in the data volume and dataset availability for analysts compels researchers to focus on data content and select high-quality datasets to enhance the performance of analytics operators. 
While selecting high-quality data significantly boosts analytical accuracy and efficiency, the exact process is very challenging given large-scale dataset availability. 
To address this issue, we propose a novel methodology that infers the outcome of analytics operators by creating a model from the available datasets. 
Each dataset is transformed to a vector embedding representation generated by our proposed deep learning model \emph{NumTabData2Vec}, where similarity search are employed. 
Through experimental evaluation, we compare the prediction performance and the execution time of our framework to another state-of-the-art modelling operator framework, illustrating that our approach predicts analytics outcomes accurately, and increases speedup. 
Furthermore, our vectorization model can project different real-world scenarios to a lower vector embedding representation accurately and distinguish them.
\vspace{-0.1in}
\end{abstract}
\vspace{-0.45in}
\section{Introduction}
\vspace{-0.15in}
Big data technologies daily face the rapid evolution in volume as well as variety and velocity of processed data \cite{b1Intdata_quality_for_data_scienceBenjaminHazen}. Such big data characteristics routinely force analytics pipelines to underperform, requiring continuous maintenance and optimization. One major reason for this is bad data quality\footnote{\href{https://tinyurl.com/de62sf48}{https://tinyurl.com/de62sf48}}. Poor data quality leads to low data utilisation efficiency and even brings forth serious decision-making errors\cite{b2DataQualityBigDataChai}.

Data quality can be improved when focusing on the actual content of the data. Data-centric Artificial Intelligence (AI) \cite{b3IntDataCentricAI} emphasises on the quality, context, and structure of the data to improve its quality, as well as the analytical or machine learning (ML) algorithmic performance. Understanding the data context properties, such as data features, origins, relevance, and potential biases, plays a critical role in modelling more accurate and reliable models. Data-centric AI prioritises the process of refining and enriching datasets to make them more suitable for real-world applications. Similarly, many researchers argue that prioritizing content-focused data quality is essential for achieving superior results \cite{b3IntDataCentricAI}.

Yet, the plethora of available data sources and datasets in an organisation data repository poses a significant challenge: Deciding the most suitable datasets for analytics workflows to ensure accurate results/predictions. While modern analytics workflows incorporate diverse operators, optimising dataset selection using data-centric AI methods remains an active research area \cite{b4IntDataCentricAI2}. When dataset selection is left to human experts, prediction performance drops, and it consumes more time. Equally costly and inefficient is the evaluation of all available datasets to identify high-quality inputs.

In previous work \cite{b7Apollo1}, predicting the output of an analytics operator assuming a plethora of available input datasets was tackled via the creation of an all-pair similarity matrix, which, relative to the similarity function used, reflected the distance between datasets over a single data quality metric (e.g., data distribution). Data or vector embeddings have been proposed to enhance big data analysis and modern AI systems. Data embedding vectorization \cite{b8Word2Vec, b9Graph2Vec} aims at projecting data from a high-dimensional representation space into a more compact, lower-dimensional space. Extracting meaningful information through data features using deep learning, data is projected to a lower representation space.  

To improve the accuracy of a modelled analytic operator (i.e., predict the outcome of a ML algorithm without actually executing it due to its cost), we propose a framework that uses vector embeddings for dataset selection from a large data lake repository. Our method predicts an operator's output for an ``unseen'' query dataset,  by selecting \emph{qualitatively similar} datasets through similarity search over the vector embeddings. The selection of similar datasets reduces the prediction error, as well as the cost to model the operator, under the assumption that realistic analytical operators perform similarly under similar inputs. The embeddings are generated using our deep learning method, \textit{NumTabData2Vec}, which processes entire tabular datasets rather than chunks or metadata, enabling efficient distinction between datasets and flexible modelling of multiple operators. Compared to similar previous work \cite{b7Apollo1, b7Apollo2}, our work uses state-of-the-art data representation (vector embeddings) which are able to capture multiple data properties that can be used in order to assess similarity, namely record order, dataset size, data distribution, etc.

The main contributions of our work can be summarised as follows:
\vspace{-0.1in}
\begin{itemize}
\item We introduce a framework for operator modelling (open-source prototype\footnote{\href{https://github.com/aloizo03/VEnOM-A-Vector-Embedding-Operator-Modelling-Framework-}{GitHub Repository}})  in order to predict its outcome on an unseen tabular input dataset from a plethora of available ones. Our method uses dataset vector embedding representations to improve the prediction performance via selecting the most relevant datasets to base its prediction upon.
\item We develop a deep learning model architecture that transforms an entire tabular dataset of numerical values to a vector embedding representation.
\item We provide an experimental evaluation of our proposed methodology using multiple real-world scenarios and compare it directly to the Apollo system \cite{b7Apollo1, b7Apollo2}.
\vspace{-0.1in}
\end{itemize}
Our evaluation illustrates that our methodology produces low prediction error by adaptively selecting similar quality datasets, achieving significant amortized speed-ups. \textit{NumTabData2Vec} evaluation shows that it effectively projects datasets into vector embeddings while accurately capturing diverse dataset properties within the representation space.

\vspace{-0.2in}
\section{Related Work}
\vspace{-0.1in}
Prior efforts have focused on boosting algorithm performance by increasing data input (record number) rather than assessing quality. Consequently, we review works that identify optimal data features for analytic operator optimization. Vectorising data to lower embedding representation is a modern method that helps in identifying significant features across data types and datasets. As vector embeddings extract important features from data, we discuss studies that used the feature representation of data tuples to improve ML model prediction.

\vspace{-0.2in}
\subsection{Data Quality}
\vspace{-0.1in}
Big data applications aim to improve data quality by addressing various challenges. Dagger \cite{b4DaggerRW} enhances data quality by detecting pipeline errors using an SQL-like language, while ReClean \cite{b4ReClean} automates tabular data cleaning via reinforcement learning. IterClean \cite{b5IterCleanRW} employs a large language model (LLM) to iteratively clean data by labelling initial tuples and using error detection, verification, and repair. In \cite{b6DataShapley}, data tuple quality is measured using Shapley values from game theory, with Truncated Monte Carlo Shapley and Gradient Shapley methods estimating a tuple's value to a learning algorithm. Apollo \cite{b7Apollo1, b7Apollo2} is a content-based method predicts analytic operator outcomes by leveraging dataset similarity through three steps: creating a similarity matrix, projecting datasets to a lower-dimensional space, and modelling the operator using a small random subset of datasets. Unlike Apollo, our approach selects the most relevant, high-quality datasets to model analytic operators, aiming to improve prediction performance. Additionally, our vector embeddings incorporate all dataset properties, whereas Apollo's \cite{b7Apollo1} similarity functions target only a single property.

\vspace{-0.2in}
\subsection{Dataset Selection Inference}
\vspace{-0.1in}

SOALA \cite{b19SOALA} selects optimal data features through online pairwise comparisons to maintain ML models over time. Its extension, Group-SOALA, introduces group maintenance to identify high-quality feature sets. In \cite{b20tfdata}, the tf.data API framework enables the creation of ML pipelines focused on selecting relevant datasets and features to improve data quality. Similarly, our framework uses dataset vector embeddings to select the most suitable datasets for modelling analytic operators or ML models, enhancing prediction accuracy.
\vspace{-0.2in}
\subsection{Data Vectorization and Embeddings}
\vspace{-0.1in}

The goal of data vectorization is to project high-dimensional data into a lower-dimensional vector space. Word2Vec \cite{b8Word2Vec} (using Continuous Bag of Words and Skip-Gram) \cite{b8Word2Vec} leverages word context to generate embeddings. Graph2Vec \cite{b9Graph2Vec} creates graph embeddings by dividing graphs into sub-graphs with a skipgram model and aggregating their embeddings. ImageDataset2Vec \cite{b11Image2Vec} extracts meta-features from image datasets to generate embeddings, helping to select the most suitable classification algorithm. Dataset2Vec \cite{b12Dataset2Vec} uses meta-features and the DeepSet model to project datasets into embeddings and measure dataset similarity. Table2Vec \cite{b13Table2Vec} generates table embeddings by incorporating data features, metadata, and structural elements like captions and column headings. Mix2Vec \cite{b24Mix2Vec}, is an unsupervised deep neural network that projects mixed data into vector embeddings. In a clustering experiment like in their work on the common Adult dataset, our model outperformed Mix2Vec (recent method without publicly available code) by nearly $10\%$, demonstrating superior performance. Inspired by these methods, we designed a model that generates vector representations of tabular datasets over their record data values, not their metadata.

Vector embeddings, which capture valuable information from data tuples, are widely used in classification tasks. TransTab \cite{b14TransTab} encodes features with transformer layers to predict classes, leveraging supervised and self-supervised pretraining. FT-Transformer \cite{b15FTransformer} and Res-Net architectures similarly use embeddings of categorical and numerical features, processed through transformer layers for class prediction, while Tab-Transformer \cite{b16TabTransformer} combines embedded categorical and normalized continuous features in an MLP for class prediction. Unlike these tuple-level approaches, our framework uses dataset-level embeddings to identify relevant datasets, enhancing analytic operator performance.

\vspace{-0.2in}
\section{Methodology}
\vspace{-0.4in}
\begin{figure}[!htbp]
    \centering
    \captionsetup{justification=centering}
    \includegraphics[width=\textwidth]{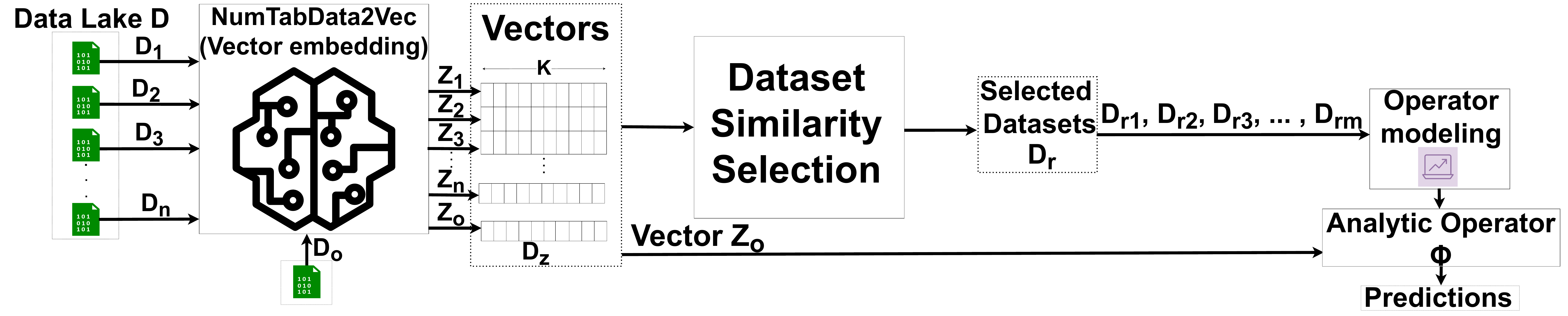}
    \vspace{-0.3in}
    \caption{Pipeline framework architecture}
    \vspace{-0.3in}
    \label{fig:fig-method1}
\end{figure}

In this section, we describe our proposed framework for modelling analytic operators over a large number of available input datasets. We also describe our approach for vectorizing tabular datasets, \emph{NumTabData2Vec}.

\vspace{-0.2in}
\subsection{Framework Architecture}
\vspace{-0.1in}
Consider a data lake repository that contains a (possibly large) number $n$ of structured tabular datasets $D = \left( d_1,\ d_2,\ d_3,\ \ldots,\ d_n\right)$. Also, let us consider an analytics operator (e.g., a ML algorithm) $\Phi$ and an ``unseen'' dataset $D_o$ (from the same domain). We assume that each $D_i, 1 \leq i \leq n $ as well as $D_o$ consist of records with numerical values only. Each dataset can, naturally, consist of different number of records. Operator $\Phi$ consumes a single such dataset as input to produce a single numerical output: $\Phi: D_i \rightarrow \mathbb{R}$. Our goal is to predict $\Phi(D_o)$ with minimal cost and error by modelling the operator's output for $D_o$ using a small subset of similar datasets $D_r \subseteq D$.

Datasets in $D_r$ closely match $D_o$ in their properties (e.g., order, distribution, and size to name a few). Previous work \cite{b7Apollo1, b7Apollo2} had to use separate similarity functions for each such property. In contrast, we leverage the embedding vectorization ($D_z$) to efficiently identify the most similar datasets using all dataset properties.

Figure \ref{fig:fig-method1} depicts the pipeline of our proposed framework. Datasets in $D$ are transformed into $k$-dimensional vectors using our \textit{NumTabData2Vec} scheme, and these embeddings are stored for reuse. Each time a $D_o$ needs to be inferred relative to an analytics operator $\Phi$, its vector embedding is created. The datasets used for the creation of the model are selected via similarity search to produce a small subset of relevant datasets. These chosen datasets are then used to model and predict $\Phi(D_o)$, ensuring that only high-quality, pertinent data is processed. With this approach, our framework is utilizing ``right quality''  data in its inference mechanism, with irrelevant and extraneous datasets being excluded from the modelling process.

The datasets are embedded by our \textit{NumTabData2Vec} method, which transforms each entire dataset in $D$ into a k-dimensional vector $z$ that captures all its characteristics. 
Our framework operates seamlessly across diverse real-world scenarios without modification, requiring only the specification of a repository containing distinct numerical tabular datasets. The Vector embedding $z$ is a lower-dimension representation of the dataset with the entire characteristics of the dataset being encoded. Dataset $D_o$ is similarly embedded as $z_o$. Using the embedding representation $z_o$ and applying different similarity functions over the vector representations $D_z$, we may choose the most similar subset of $D$. The final step of the pipeline involves the operator modelling with any relevant method (e.g., Linear Regression, SVM, Multi-Layer Perceptron, etc.). This model is then used in order to infer the value of $\Phi(D_o)$.
\vspace{-0.03in}

\begin{figure}[!t]
    \centering
    \vspace{-0.18in}
    \captionsetup{justification=centering}
    \includegraphics[width=0.7\textwidth]{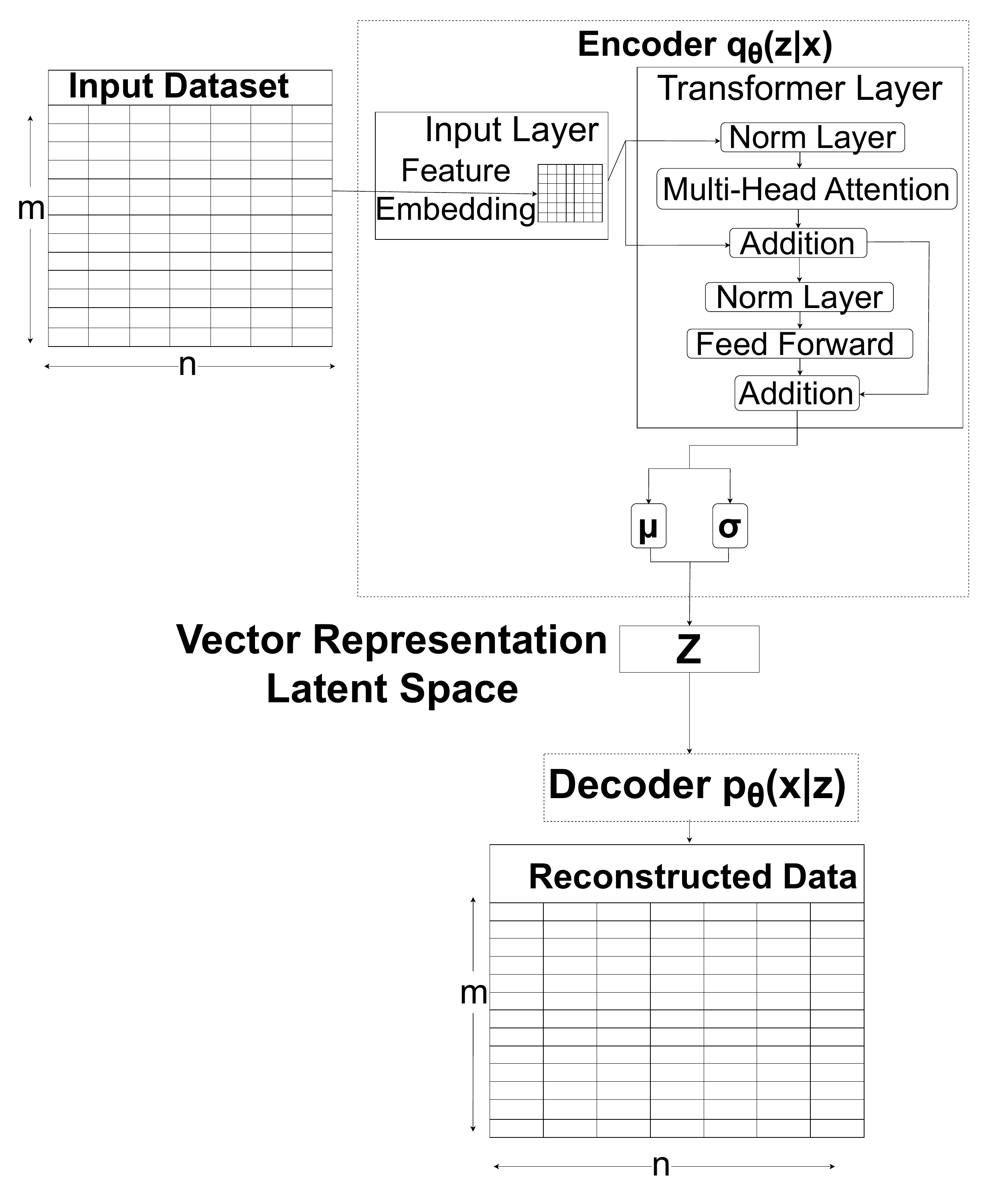}
    \vspace{-0.15in}
    \caption{NumTabData2Vec deep learning model architecture}
    \vspace{-0.3in}
    \label{fig:fig-method2}
\end{figure}

\vspace{-0.2in}
\subsection{NumTabData2Vec}
\vspace{-0.1in}

This method transforms a dataset $D_i$ into a lower-dimensional vector $z$ using only its numerical values while excluding metadata like column names and filenames. We present a deep learning model architecture based on the variational autoencoder (VAE) \cite{b1VAE} concept, that projects high-dimensional data into a $1\times k$ vector embedding dimension. The proposed method is thus defined as: 
\vspace{-0.15in}
\begin{equation}
    NumTabData2Vec\left( D_i \right) \rightarrow z, \label{eq_3_method} \\[-0.15in]
\end{equation}
where the model takes an $m\times n$ dimensional numerical dataset and projects it to a lower $k$-dimension space $z$ ($k > 1$). We desire our method to be generally applicable to any dataset by learning to project a vector embedding during training. We also expect this method to learn vector embeddings from diverse data and operate without additional training or fine-tuning. Finally, we ought our scheme to be able to quickly and precisely extract vectors from every input while handling varying dataset dimensions without modifications.  

The deep learning model architecture is depicted in Figure \ref{fig:fig-method2}, where a dataset $D_i$ with dimensions $m \times n$ passes through the encoder and is projected into a vector embedding representation $z$, then reconstructed by a decoder that mirrors the encode. The vector representation $z$ is learned using a probabilistic encoder $q_\phi\left(z|x\right)$, and decoder $p_\theta\left(x|z\right)$ that learns the distribution utilising the Kullback–Leibler (KL) divergence \cite{b2KLDivergence}. To achieve that, the following condition:
\vspace{-0.15in}
\begin{equation}
    minD_{KL}\left( q_\phi\left(z|x\right) \Vert p_\theta\left(x|z\right) \right)\label{eq_4_method} \\[-0.05in]
\end{equation} 
of Kullback–Leibler (KL) divergence must be minimised. To learn the new vector representation $z$, the dataset must be reconstructed back from $z$ to its input format using the decoder, to verify that the vector representation is compact. This reconstruction loss is part of the overall loss function, defined as:
\vspace{-0.1in}
\begin{equation}
     \resizebox{.85\hsize}{!}{ $\mathcal{L}_{\theta, \phi}\left(x\right) = \mathbb{E}_{q_\phi\left(z|x\right)}\left(\log{\left(p_\theta\left(x|z\right)\right)}\right) - D_{KL}\left( q_\phi\left(z|x\right) \Vert p_\theta\left(x|z\right) \right) \label{eq_5_method} $} \\[-0.05in]
\end{equation}
This loss function is called evidences lower bound (ELBO). While the KL divergence is minimised to learn the vector embedding representation $z$, the ELBO is maximized so the condition, 
\vspace{-0.15in}
\begin{equation}
      argmax\mathcal{L}_{\theta, \phi}\left(x\right) \label{eq_6_method} \\[-0.05in]
\end{equation}
must be satisfied. The Decoder $p_\theta\left(x|z\right)$ is only used during the training phase to teach the encoder how to project the vector embedding representation $z$.

The decoder extracts feature embeddings from the input dataset $D_i$ and processes them through Transformer layers. For the transformer layer we are using the pre-LN Transformer layer \cite{b3PreNormTL} instead of the traditional post-LN Transformer layer where the normalisation layer is employed inside the residual connection and before the prediction of the feed-forward layer. Following that, the transformed embedding space is projected into a probabilistic vector space z using the mean ($\mu$) and standard deviation ($\sigma$). This lower-dimensional space retains all essential information about $D_i$, and a higher dimension $k$ in $z$ leads to a more accurate representation by capturing additional features \cite{b12Dataset2Vec}.

\vspace{-0.2in}
\subsection{Dataset Selection}
\vspace{-0.1in}

\begin{algorithm}[t]
\caption{K-Means Clustering Algorithm for dataset selection}
\label{algo_1_K_Means}
\begin{algorithmic}[1]
\Require Vectors $\mathbf{Z} = \{\mathbf{z}_1, \mathbf{z}_2, \dots, \mathbf{z}_n\}$, $D_o$ vector $z_o$, range of clusters $S = (2, \dots, p)$, Maximum size of cluster $max_s$, Minimum size of cluster $min_s$
\State \textbf{Initialize} the number of cluster $s$ using Silhouette score: $s = Silhouette Score(Z, S)$
\State \textbf{Initialize} the $s$ cluster centroids $\mathbf{C} = \{\mathbf{c}_1, \mathbf{c}_2, \dots, \mathbf{c}_s\}$ randomly from the vectors $\mathbf{Z}$.
\Repeat
    \State \textbf{Assignment Step:}
    \For{each vector $\mathbf{z}_i \in \mathbf{Z}$}
        \State Assign $\mathbf{z}_i$ to the nearest centroid based on Euclidean distance:
        \vspace{-0.1in}
        \[
        \text{Assign } \mathbf{z}_i \text{ to cluster } j = \arg\min_{j} \|\mathbf{z}_i - \mathbf{z}_j\|^2
        \]
        \vspace{-0.25in}
    \EndFor
    \State \textbf{Update Step:}
    \For{each centroid $\mathbf{c}_j$}
        \State Update $\mathbf{c}_j$ as the mean of all vectors assigned to cluster $j$:
        \vspace{-0.13in}
        \[
        \mathbf{c}_j = \frac{1}{|\{ \mathbf{z}_i \in C_j \}|} \sum_{\mathbf{z}_i \in C_j} \mathbf{v}_i
        \]
         \vspace{-0.22in}
    \EndFor
\Until{Centroids $\mathbf{C}$ do not change significantly}
\State Save the cluster model \textbf{K-Means}
\State Find in which cluster the vector $z$ belongs, $c = \textbf{K-Means}(z)$
\State Find which datasets $D_r$ are belongs to cluster $c$
\State Check the number of datasets in $D_r$ and update it if it does not meet the $min_s$ and $max_s$.
\State \textbf{Return} Datasets $D_r$
\end{algorithmic}
\end{algorithm}

The selection of the most similar datasets has been implemented using three different approaches. Different similarity functions are easily plugged into our pipeline. The first method uses cosine similarity, which measures the angle between two vectors in the embedding space $z$ independent of their magnitudes, with a higher value indicating greater similarity. 
The alternative method calculates the Euclidean distance between two vectors to capture their geometric closeness and determine dataset similarity. This approach aids in selecting the most relevant datasets according to organizational requirements. 
The smaller the distance value then more similar are the datasets. Dataset selection using cosine similarity or euclidean distance selects a fraction of $\lambda, \lambda > 0$ of the closest datasets to dataset $D_o$. The third approach involves utilising the K-Means \cite{b17KMeansa, b17KMeansb} clustering technique to choose the most relevant datasets. The datasets from $D$ are divided into $s$ ($s > 1$) separate clusters, where datasets with similar features are grouped together in the same cluster based on similarity equations.
We determine the optimal number of clusters using silhouette scores \cite{b18Silhouettes}. This is done by the following equation:
\vspace{-0.1in}
\begin{equation}
    SilhouetteScore(z_i) = \frac{b(z_i) - a(z_i)}{\max(a(z_i), b(z_i))} \label{eq_9_shilouete_score}, \\[-0.1in]
\end{equation}
where for each vector $z_i$ computes the mean intra-cluster distance ($a(z_i)$) which is the distance with the other vectors in the same cluster, and the mean nearest-cluster distance ($b(z_i)$) is the minimum average distance with the other vectors in a different cluster. Silhouette Score ranges from $-1$ to $1$, and the higher value defines the best $s$ number for clusters.

Algorithm \ref{algo_1_K_Means}, outlines the K-means process for selecting relevant datasets $D_r$ based on the target dataset $D_o$. Using the optimal s (with the highest Silhouette score), the vector representations $z$ of each dataset are clustered. Next, the algorithm uses the vector $z_o$ of dataset $D_o$ to find the closest cluster centroid. All datasets in that cluster are defined as the relevant datasets $D_r$, which are then used to model the analytics operator. However, we defined a maximum and minimum size for $D_r$, and if these conditions are not met, datasets are either removed from the cluster or added based on their distance from the cluster centroid. These small adjustments are only made in cases where the clustering technique does not yield results that satisfy our requirements.

 \vspace{-0.2in}
\section{Evaluation}
 \vspace{-0.15in}
We compared our framework with Apollo \cite{b7Apollo1, b7Apollo2}, which models analytic operators using data content. Two loss functions to measure prediction accuracy are employed: root-mean-square error (RMSE) and mean absolute error (MAE). RMSE is sensitive to outliers, while MAE is not; conversely, RMSE accounts for error direction, which MAE cannot. We further assess efficiency using \textit{Speedup} and \textit{Amortized Speedup} metrics, where \textit{Speedup} is defined as $\frac{T{^{(i)}_{op}}}{T{^{(i)}_{SimOp} + T_{vec} + T_{sim} + T_{pred}}}$, where $T{^{(i)}_{op}}$ is the time to execute operator $i$ on all datasets, $T{^{(i)}_{SimOp}}$ is the time to model the operator with datasets from similarity search, $T_{vec}$ is the vector embedding computation time, $T_{sim}$ is the similarity search time, and $T_{pred}$ is the prediction time for $D_o$. Amortized speedup including one-time vectorization per data lake across multiple operators. Three variants with vector sizes 100, 200, and 300 (each with eight transformer layers) were trained for $100$ epochs on four NVIDIA A$10$ GPUs. More experimental evaluation results can be found in the extended version of this work \cite{loizou2025data}.
 \vspace{-0.2in}
\subsection{Evaluation Setup}
 \vspace{-0.1in}
Our framework is deployed over an AWS EC2 virtual machine server running with 48 vCPUs of AMD EPYC 7R32 processors at 2.40GHz, and four A10s GPUs with 24GB of memory each, $192GB$ of RAM memory, and $2TB$ of storage, running over Ubuntu 24.4 LTS. Our code is written in Python (v.3.9.1) and PyTorch modules (v.2.4.0). Apollo was deployed on the same AWS EC2 virtual machine server, utilizing only the vCPUs and RAM, as it does not require a GPU for execution. 

 \vspace{-0.2in}
\subsection{Datasets}
 
\begin{table}[!ht]
\vspace{-0.5in}
    \centering
    \setlength\doublerulesep{0.5pt}
    \caption{Dataset properties for experimental evaluation}
    \label{tab:table-evaluation-datasets}
    \scalebox{0.7}{
    \begin{tabular}{||c|c|c|c||}
        \hline
         \makecell{Dataset Name}& \makecell{\# Files} & \makecell{\# Tuples} & \makecell{\# Columns}\\ \hline\hline
         Household Power & & & \\
         Consumption \cite{b21HPCdataset} & $401$ & $2051$ & 7\\
         \hline
         Adult \cite{b22AdultDataset} & $100$ & $228$ & 14\\
         \hline
         Stocks \cite{b23StockMarketDataset} & $508$ & $1959 - 13$ & 7 \\
         \hline
         Weather \cite{b23WeatherDataset} & $49$ & $516$ & 7 \\ \hline
    \end{tabular}
    }
\vspace{-0.2in}
\end{table}
 \vspace{-0.15in}
We evaluated our framework using four real-world datasets (see Table \ref{tab:table-evaluation-datasets}). The NumTabData2Vec module was trained on separate data ($60\%$  training, $40\%$ testing). The Household Power Consumption (HPC) dataset \cite{b21HPCdataset} contains $401$ datasets with $2051$ tuples and seven features recorded at one-minute intervals of electric power usage measurements. The Adult dataset \cite{b22AdultDataset}, used for binary classification, predicts income levels and includes $100$ datasets with $228$ individuals and socio-economic features. The Stock Market dataset \cite{b23StockMarketDataset} consists of $508$ datasets with $13$ to $1959$ tuples describing daily NASDAQ stock prices. Weather dataset \cite{b23WeatherDataset} provides hourly measurements from $36$ U.S. cities (2012–2017), split into $49$ datasets with $516$ tuples and seven features. 
Any categorical feature column in all datasets is transformed to numerical data by one-hot encoding. 
These datasets were selected to demonstrate our framework's ability to perform consistently across diverse real-world scenarios.

Our framework was evaluated by predicting the outputs of various ML operators without directly executing them. Datasets were projected into $k$-dimensional spaces with vector dimensions of $100$, $200$, and $300$. For regression datasets (Household Power Consumption and Stock Market), we modelled Linear Regression (LR) and Multi-Layer Perceptron (MLP), while for classification datasets (Weather and Adult), we modelled Support Vector Machine (SVM) and MLP classifiers. Each experiment has executed 10 times, and we report the average error loss and speedup. 
 \vspace{-0.2in}
\subsection{Evaluation Results}
 \vspace{-0.1in}

\begin{figure*}[!t]
    \vspace{-0.15in}
    \begin{minipage}{0.48\textwidth} 
        \subfloat[Linear Regression RMSE error loss]{
        \includegraphics[width=0.48\textwidth]{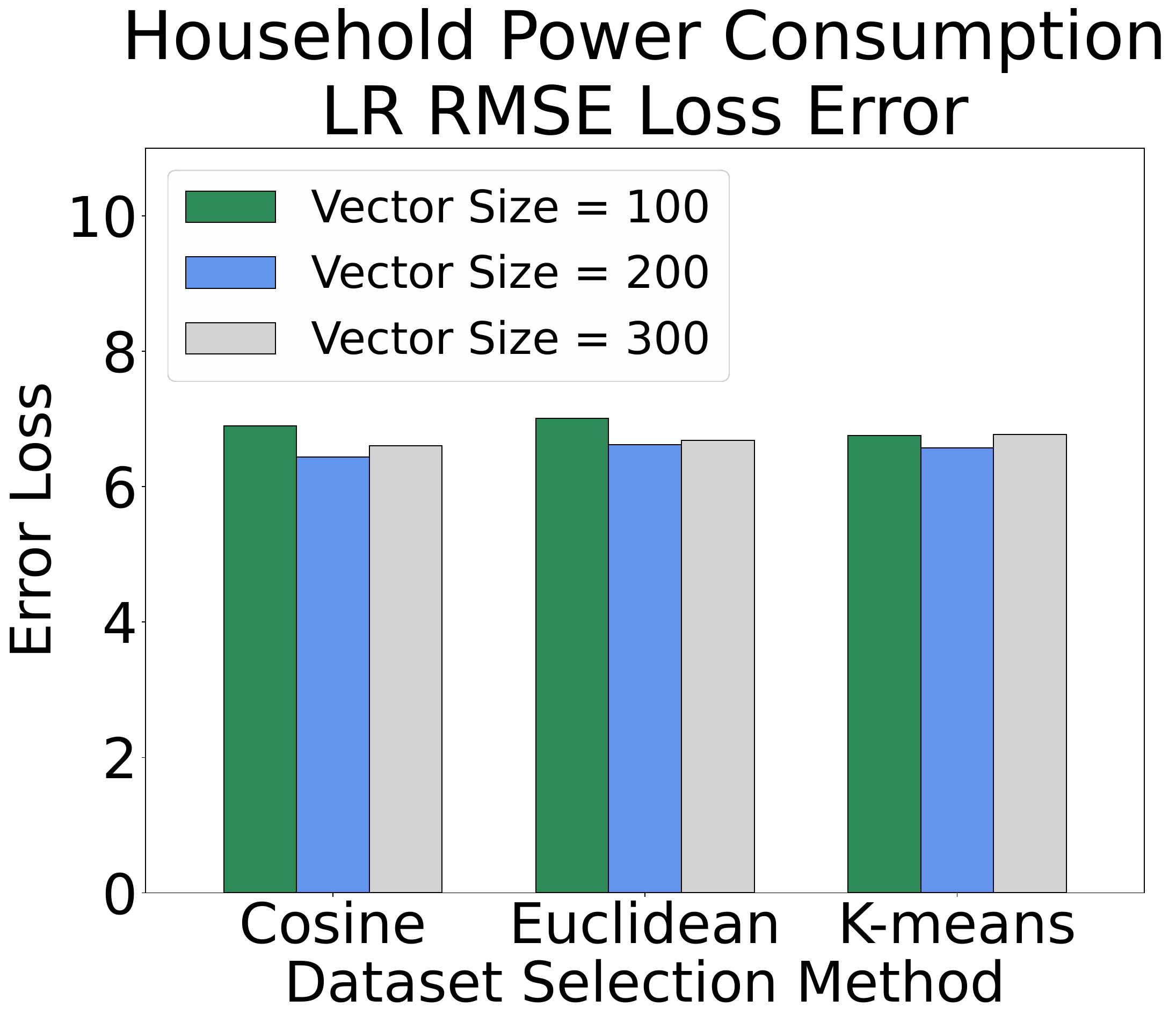}
        \label{fig:HPC-LR-RMSE}
    }
    \subfloat[Linear Regression MAE error loss]{
        \includegraphics[width=0.45\textwidth]{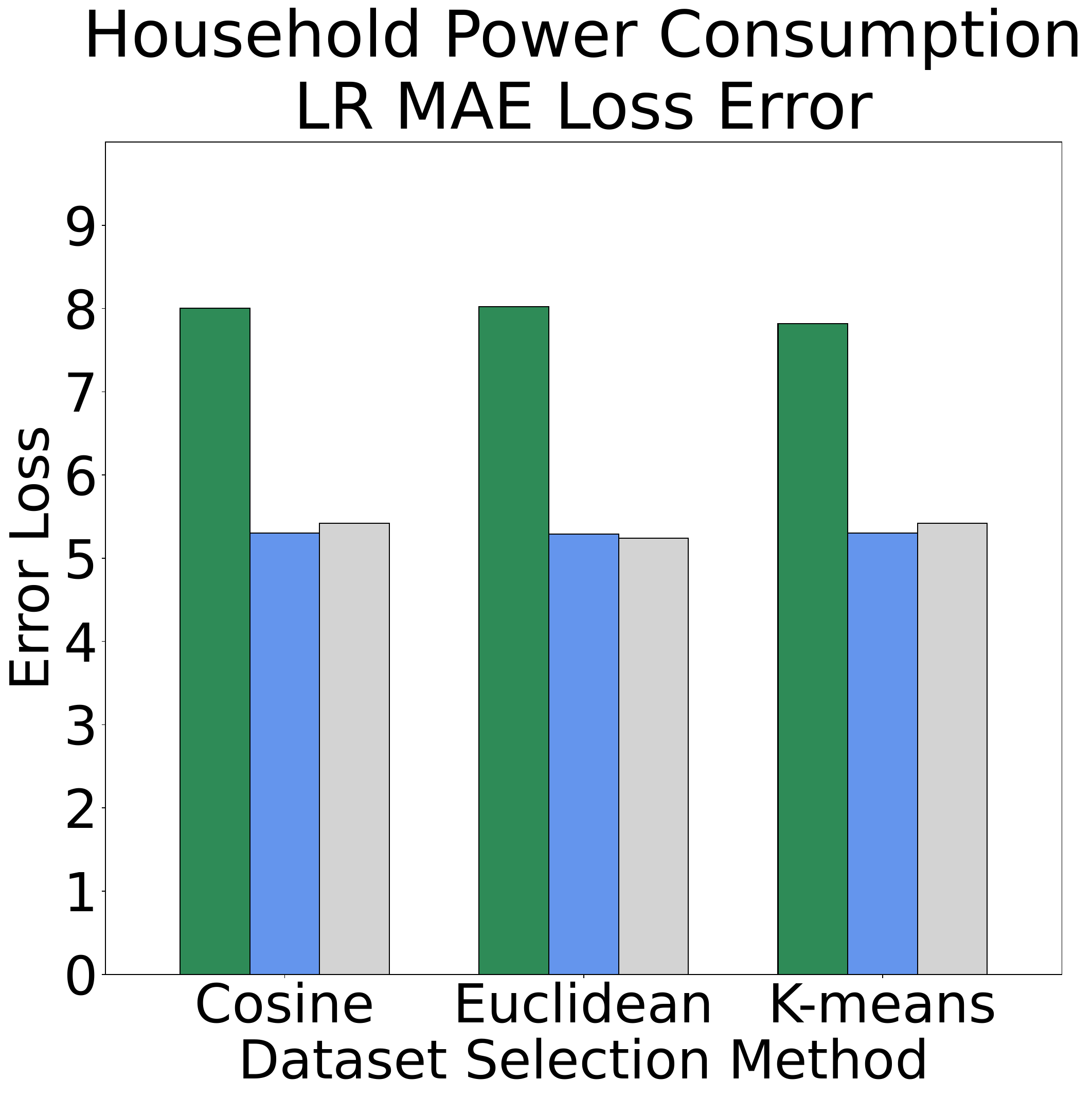}
        \label{fig:HPC-LR-MAE}
    }
    
    \subfloat[MLP for Regression RMSE error loss]{
        \includegraphics[width=0.48\textwidth]{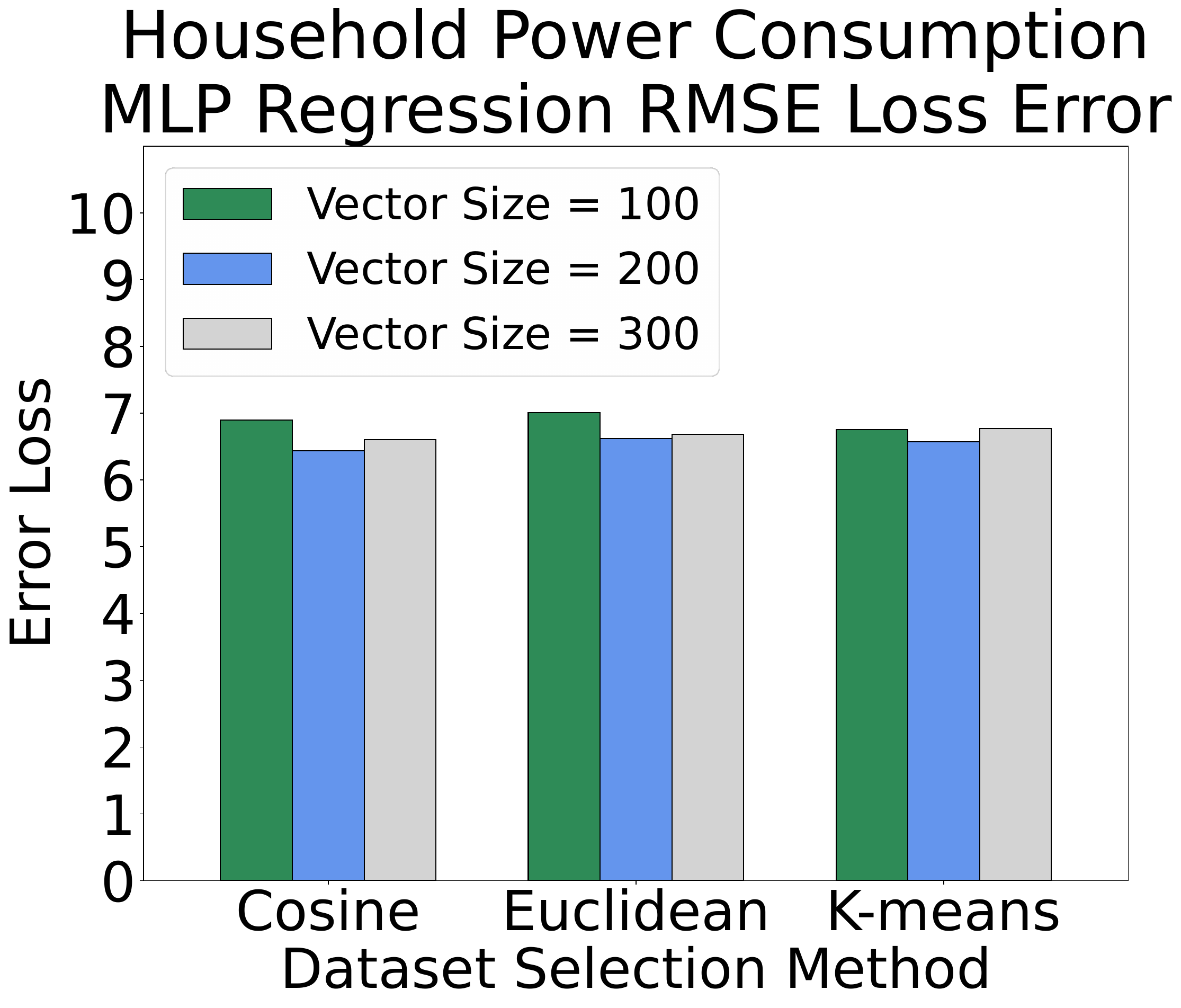}
         \label{fig:HPC-MLP-RMSE}
    }
    \subfloat[MLP for Regression MAE error loss]{
         \includegraphics[width=0.48\textwidth]{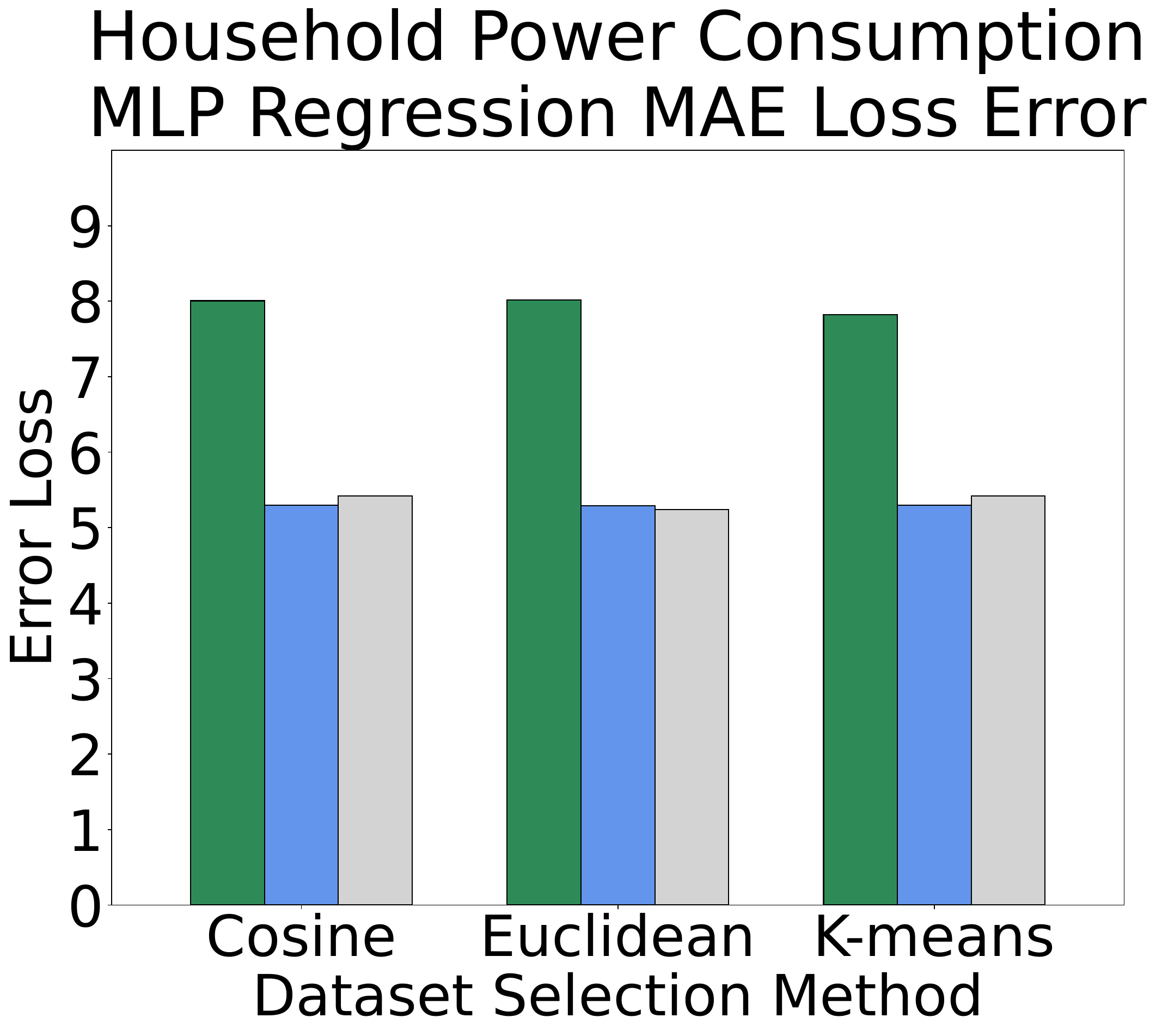}
         \label{fig:HPC-MLP-MAE}
    }
    \vspace{-0.15in}
    \caption{Household power consumption dataset prediction error loss}
    \vspace{-0.33in}
    \label{fig:HPC-EVAL-RES}
    \end{minipage}
    \hfill
    \begin{minipage}{0.48\textwidth} 
        \centering
     \subfloat[Linear Regression RMSE error loss]{
        \includegraphics[width=0.49\textwidth]{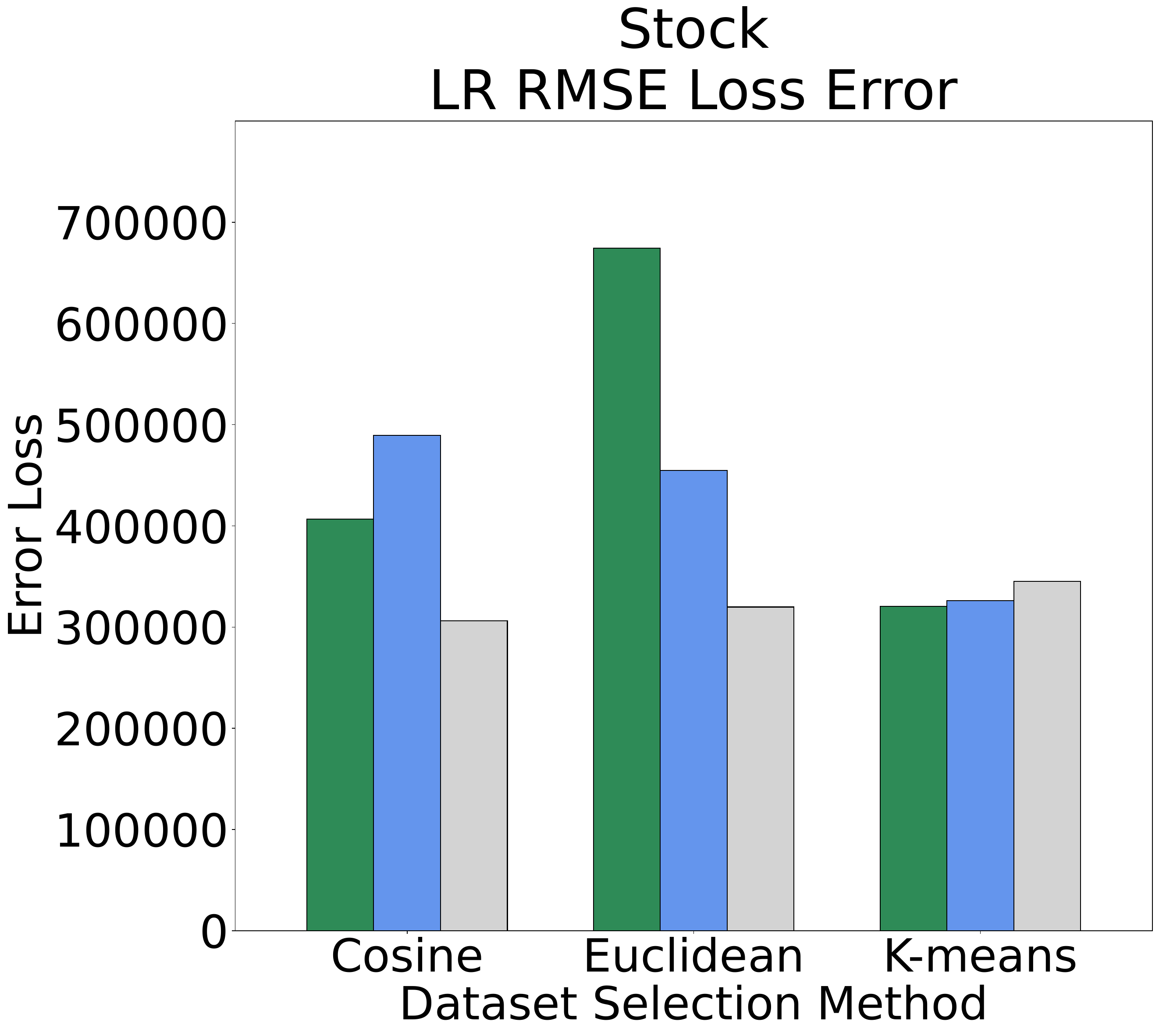}
         \label{fig:Stock-LR-RMSE}}
    \subfloat[Linear Regression MAE error loss]{
         \includegraphics[width=0.49\textwidth]{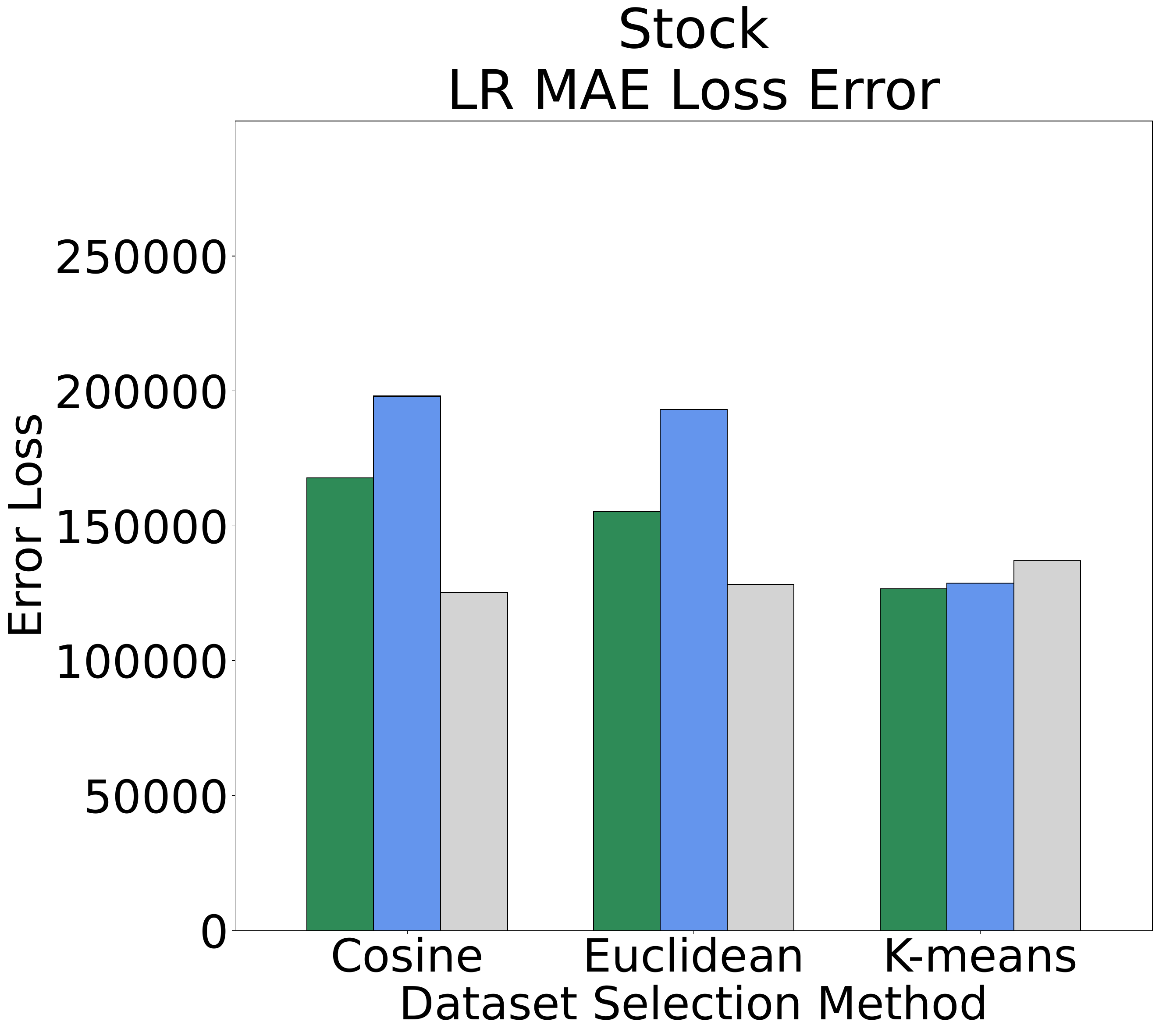}
        \label{fig:Stock-LR-MAE}}
     
     \subfloat[MLP for Regression RMSE error loss]{
         \includegraphics[width=0.49\textwidth]{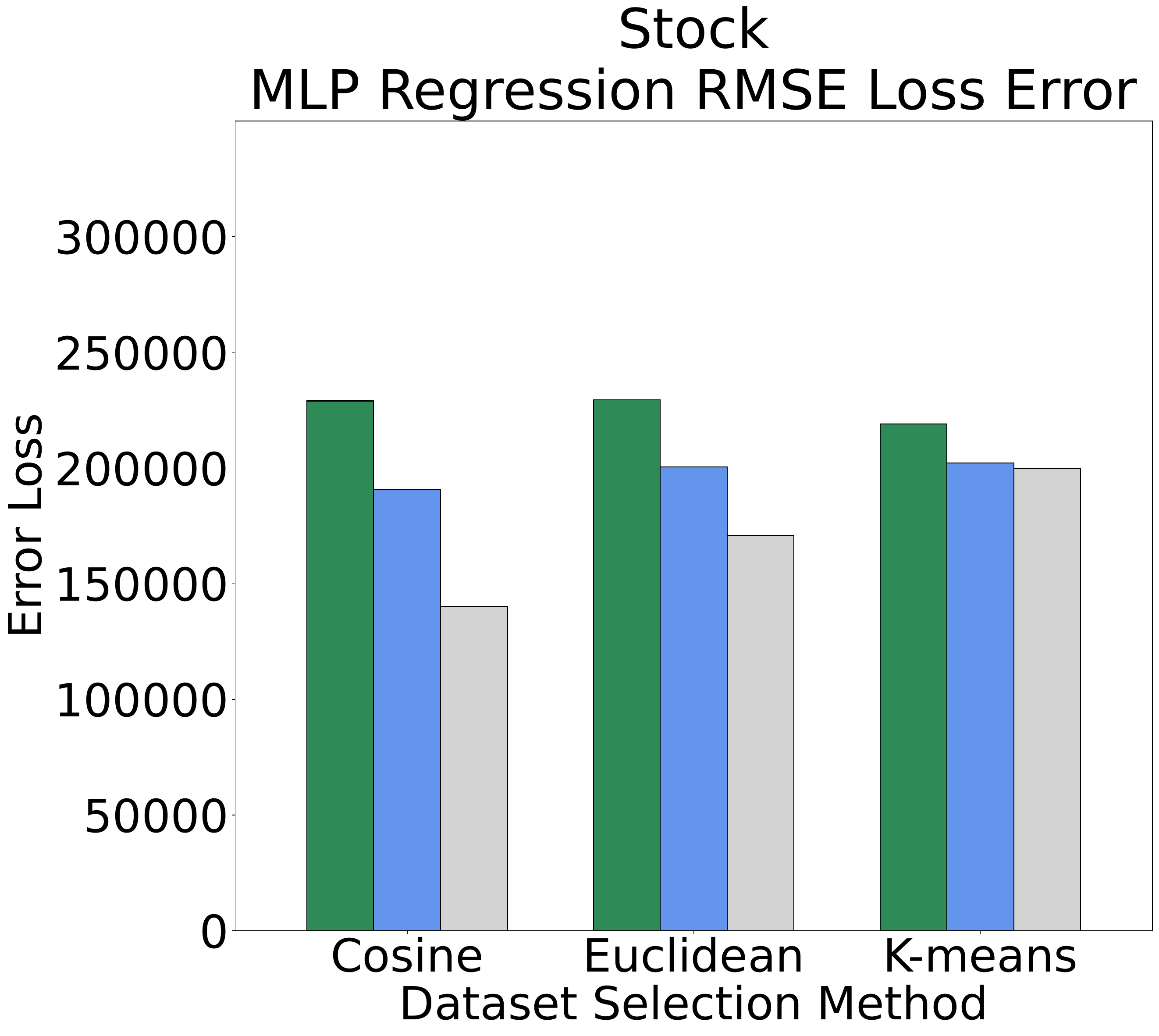}
         \label{fig:Stock-MLP-RMSE}}
    \subfloat[MLP for Regression MAE error loss]{
         \includegraphics[width=0.49\textwidth]{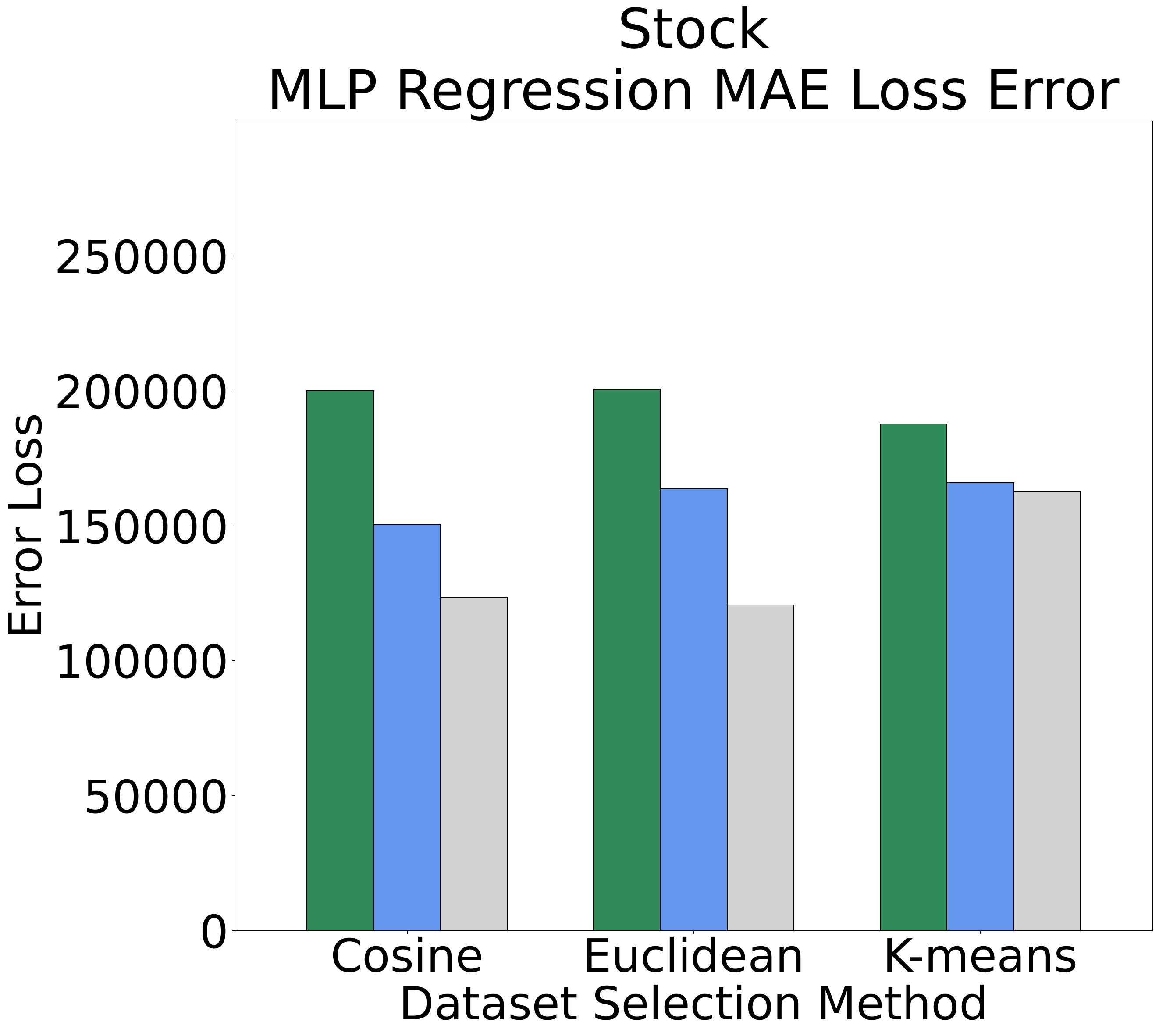}
         \label{fig:Stock-MLP-MAE}
    }
    \vspace{-0.15in}
        \caption{Stock market dataset prediction error loss}
        \vspace{-0.33in}
        \label{fig:Stock-EVAL-RES}
    \end{minipage}
\end{figure*}

\begin{figure*}[!htpb]
\vspace{-0.2in}
    \begin{minipage}{0.48\textwidth} 
       \subfloat[SVM with SGD RMSE error loss]{
        \includegraphics[width=0.48\textwidth]{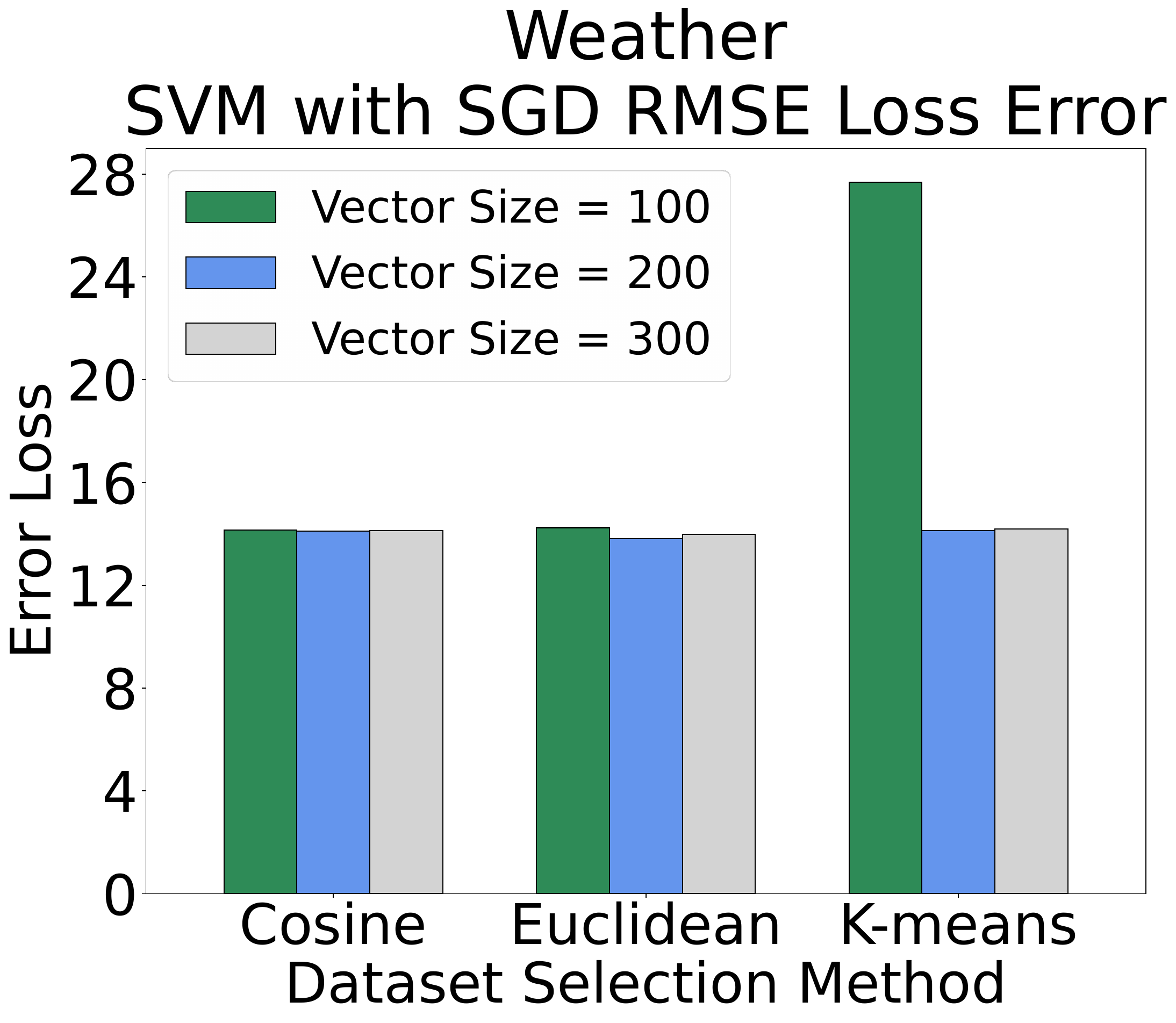}
         \label{fig:Weather-SVM-RMSE}
    }
    \subfloat[SVM with SGD MAE error loss]{
        \includegraphics[width=0.48\textwidth]{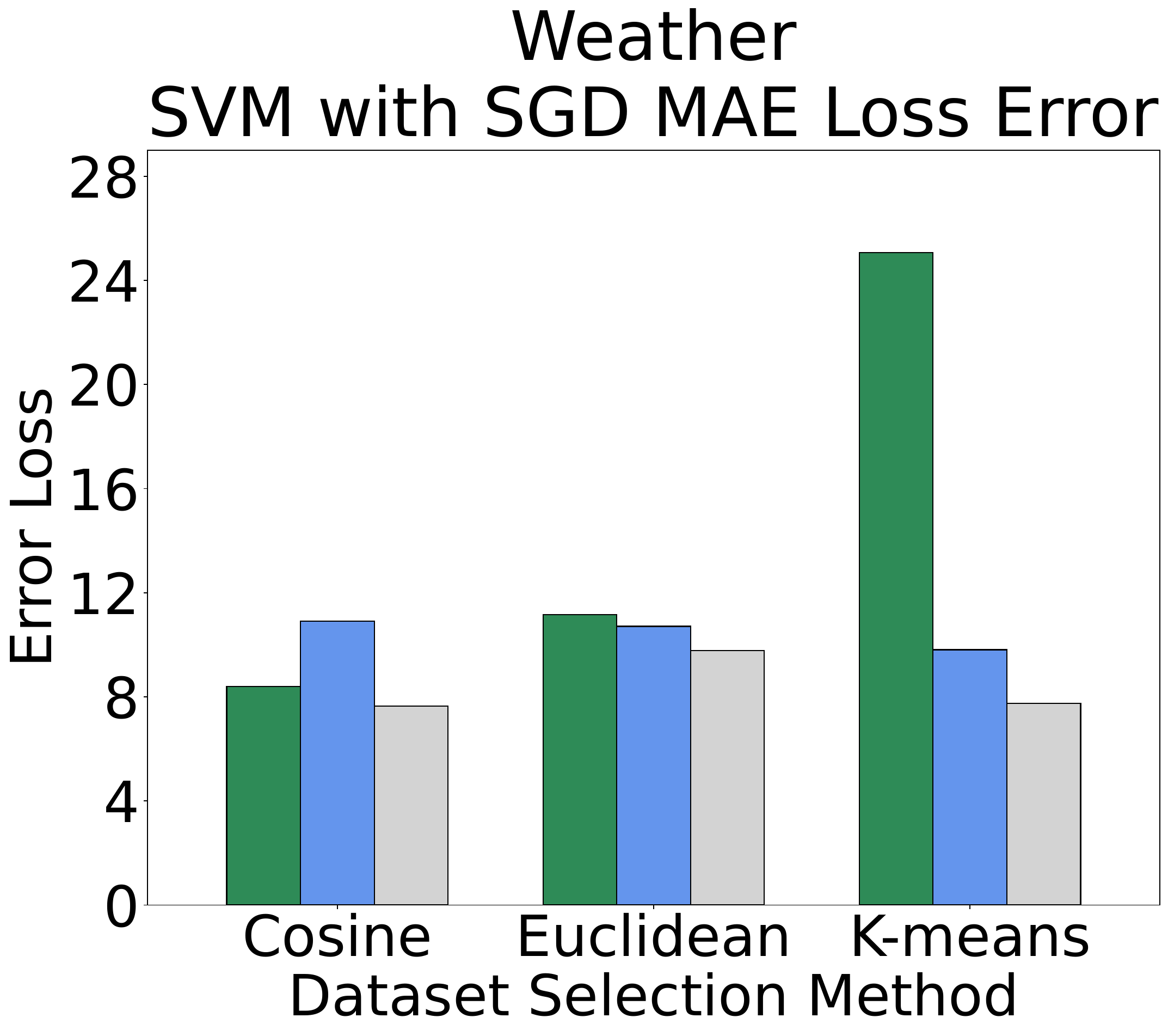}
         \label{fig:Weather-SVM-MAE}
    }
     
     \subfloat[MLP RMSE error loss]{
         \includegraphics[width=0.48\textwidth]{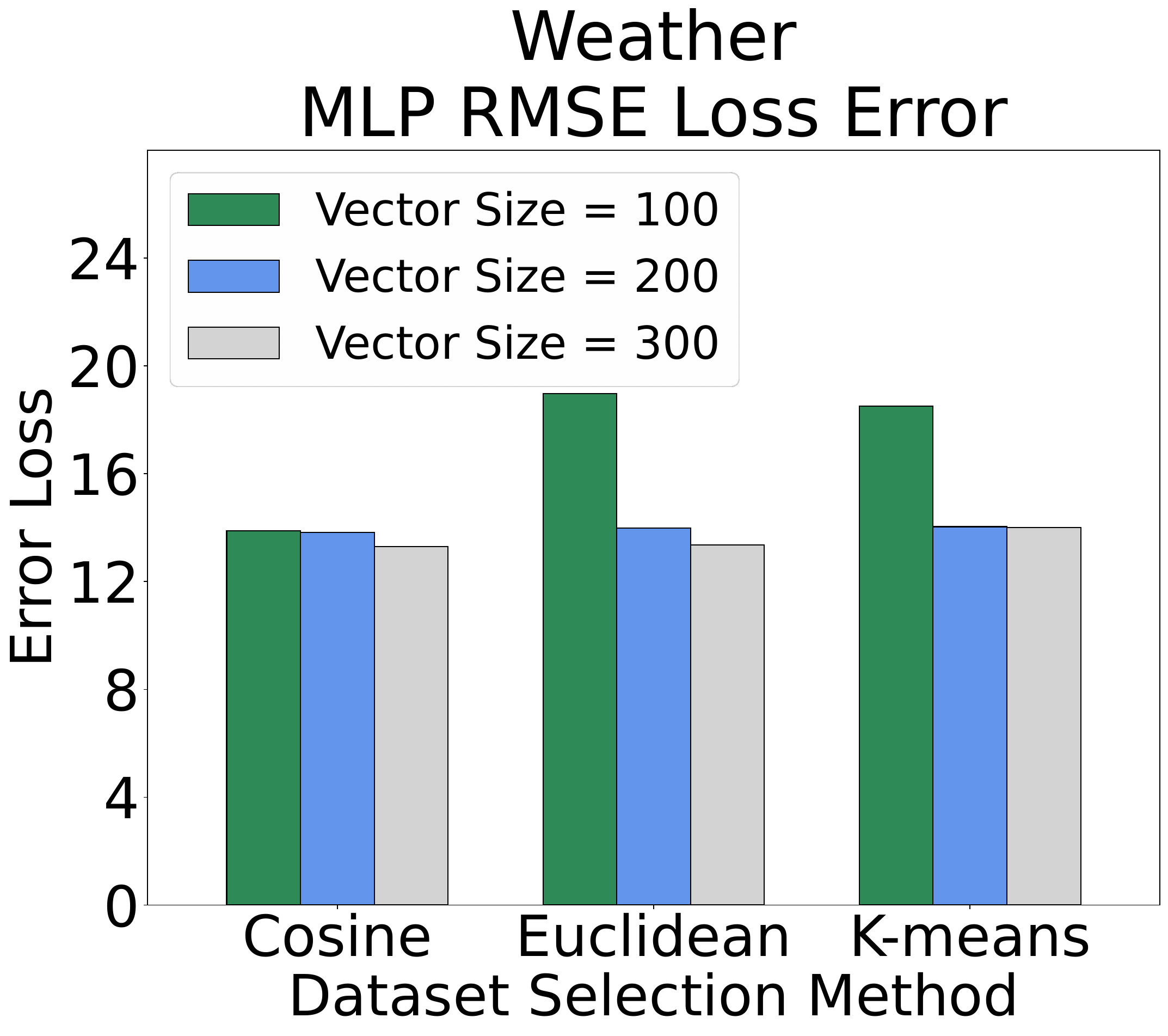}
         \label{fig:Weather-MLP-RMSE}
    }
    \subfloat[MLP MAE error loss]{
         \includegraphics[width=0.48\textwidth]{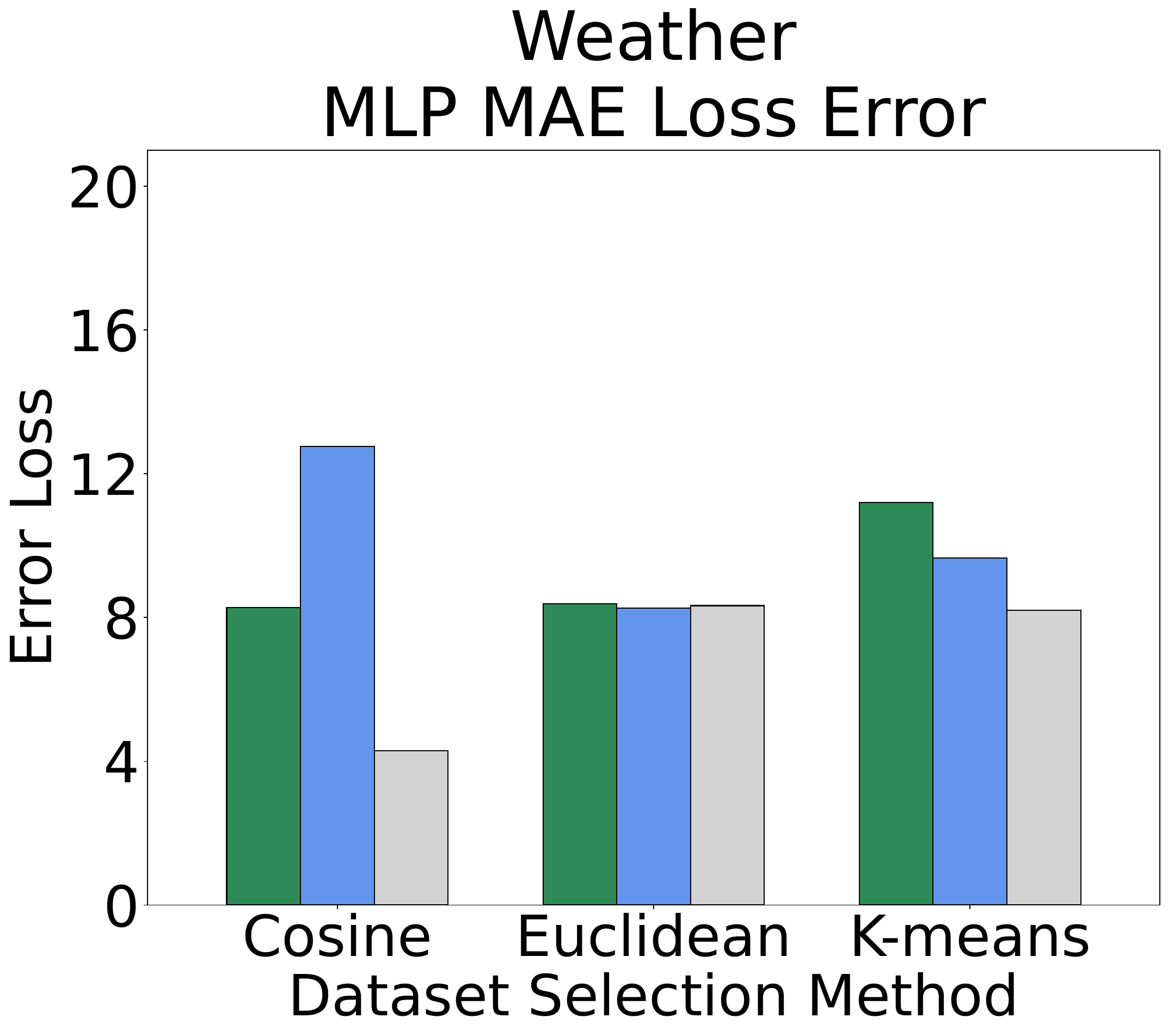}
         \label{fig:Weather-MLP-MAE}
    }
        \vspace{-0.18in}
        \caption{Weather dataset prediction error loss}
        \vspace{-0.4in}
        \label{fig:Weather-EVAL-RES}
    \end{minipage}
    \hfill
    \begin{minipage}{0.48\textwidth} 
        \centering
     \subfloat[SVM with SGD RMSE error loss]{
        \includegraphics[width=0.48\textwidth]{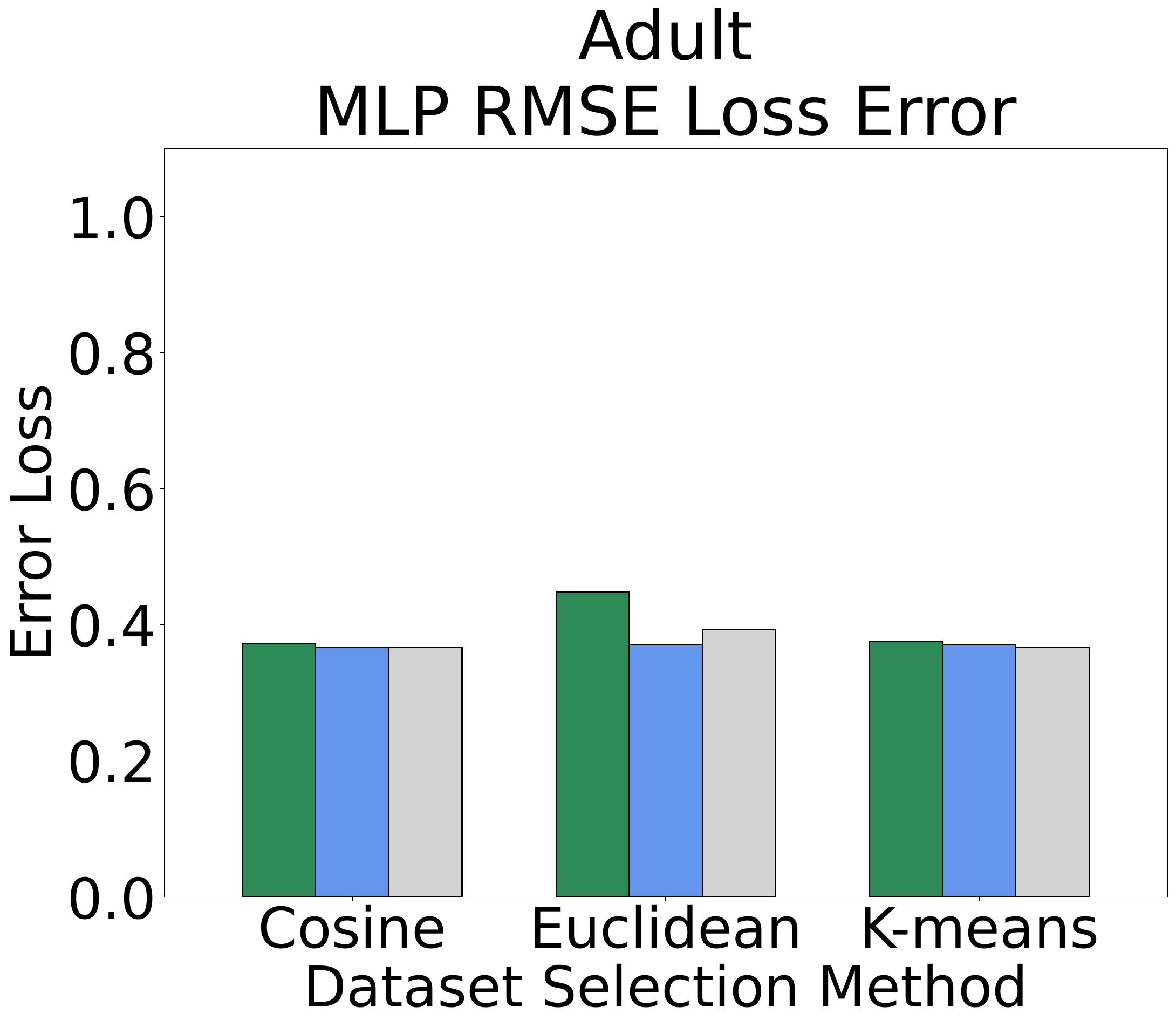}
         \label{fig:Adult-LR-RMSE}
    }
    \subfloat[SVM with SGD MAE error loss]{
        \includegraphics[width=0.48\textwidth]{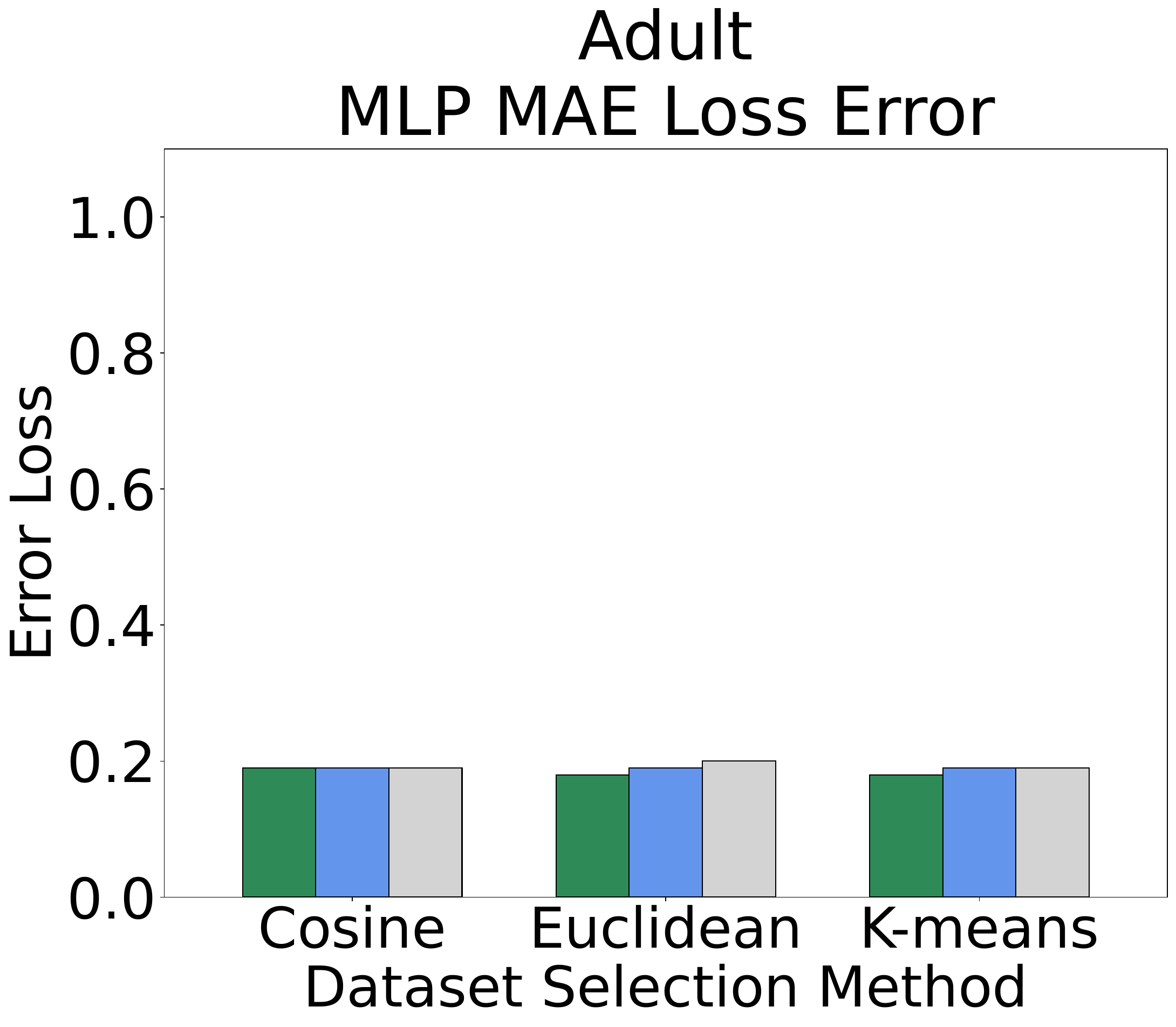}
         \label{fig:Adult-LR-MAE}
    }
     
     \subfloat[MLP RMSE error loss]{
         \includegraphics[width=0.48\textwidth]{Figures/Results/Sim_Search/Adult/Adult_MLP_RMSE_Loss_fig.pdf}
         \label{fig:Adult-MLP-RMSE}
    }
    \subfloat[MLP MAE error loss]{
         \includegraphics[width=0.48\textwidth]{Figures/Results/Sim_Search/Adult/Adult_MLP_MAE_Loss_fig.pdf}
         \label{fig:Adult-MLP-MAE}
    }
        \vspace{-0.18in}
        \caption{Adult dataset prediction error loss}
        \vspace{-0.4in}
        \label{fig:Adult-EVAL-RES}
    \end{minipage}
\end{figure*}

Figures \ref{fig:HPC-EVAL-RES}, \ref{fig:Stock-EVAL-RES}, \ref{fig:Weather-EVAL-RES}, and \ref{fig:Adult-EVAL-RES} present the evaluation results of different similarity search methods across vector embedding spaces of sizes $100$, $200$, and $300$ (green, blue, and grey bars, respectively). In each sub-figure, the y-axis represents the error loss value, while the x-axis displays the similarity search method applied over the vector embeddings. Figures \ref{fig:HPC-EVAL-RES} and \ref{fig:Stock-EVAL-RES} display results for the Stock Market and Household Power Consumption datasets, with MLP regression in the bottom sub-figure and LR in the top. Figures \ref{fig:Weather-EVAL-RES} and \ref{fig:Adult-EVAL-RES} show results for the Weather and Adult datasets, with SVM (SGD) in the top sub-figure and MLP classifier in the bottom. Left sub-figures use RMSE loss, while right sub-figures use MAE loss.

\begin{multicols}{2} 

\end{multicols}

\vspace{-0.35in}
Figure \ref{fig:HPC-EVAL-RES}, for the HPC dataset, shows as increase the vector dimension size there is slightly lower prediction error for all the operator modelling. Different similarity methods do not result in any significant differences in the prediction error loss for all the operator modelling. This suggests that, regardless the similarity selection method, our framework effectively selects the most optimal subset of data to improve model predictions on the unseen input dataset $D_o$. 
Additionally, we observe higher error loss with a vector size of 100, which can be attributed to the reduced representation capacity of lower-dimensional vectors. This limitation results in fewer ``right'' datasets being selected.

For the stock market dataset, Figure \ref{fig:Stock-EVAL-RES} depicts that a vector embedding representation of size $300$ models more accurate operators, with cosine similarity performing best in the similarity search and modelling the most optimal operator. However, due to the inherent volatility in Stock market data from different days, all models in the stock market dataset experiments exhibit high loss values. 

In the weather dataset, the SVM operator results (Figures \ref{fig:Weather-SVM-RMSE} and \ref{fig:Weather-SVM-MAE}) show that using $300$ sized vectors in the representation space consistently led to more accurate operator models across all similarity methods. Specifically, cosine similarity in combination with the $300$-dimensional vector embedding reduced the error rate in operator predictions, demonstrating that projecting datasets into this representation space and applying cosine similarity improves the prediction accuracy on the modelled operator. For the MLP classifier (Figures \ref{fig:Weather-MLP-RMSE} and \ref{fig:Weather-MLP-MAE}), the results illustrate that using vector embeddings of size $300$ and Cosine similarity- produced the most accurate MLP classifier operators.

\begin{multicols}{2} 

\end{multicols}

\vspace{-0.4in}
On the other hand, the Adult dataset shows the lowest error rates, with error loss values consistently below $0.5$ across all vector embedding dimensions and similarity search methods (see Figure \ref{fig:Adult-EVAL-RES}). The Adult dataset, besides exhibiting a high number of rows, also has a higher number of columns, which demonstrates that our framework performs consistently well even with larger datasets.
Additionally, we observe that the lowest prediction error across all datasets occurs when using higher-dimensional vector embeddings. With a trade-off between accuracy and execution time as the difference to generate all data lake available datasets vector embedding representation between $100$ and $300$ size dimension in the vector representation space to be less than $60$ seconds. This confirms that a higher number of vector dimensions leads to more accurate predictions, consistent with findings in previous research \cite{b8Word2Vec}.

\begin{figure}[!hb]
     \centering
     \vspace{-0.3in}
     \begin{subfigure}[b]{0.47\textwidth}
         \centering
         \includegraphics[width=\textwidth]{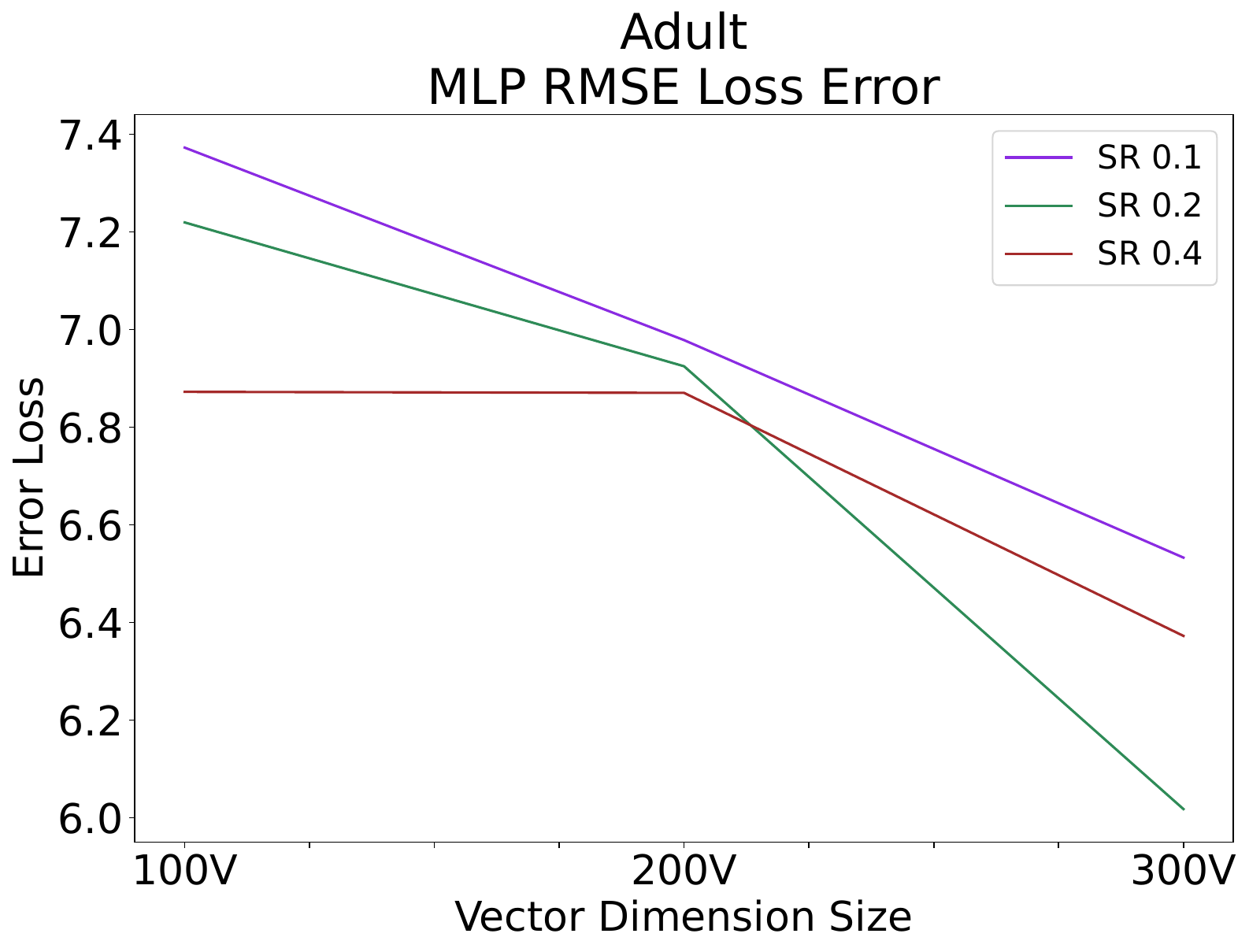}
         \caption{Adult Dataset MLP Operator RMSE error loss}
         \label{fig:Adult-SR-RMSE}
     \end{subfigure}
     \hfill 
     \begin{subfigure}[b]{0.49\textwidth}
         \centering
         \includegraphics[width=\textwidth]{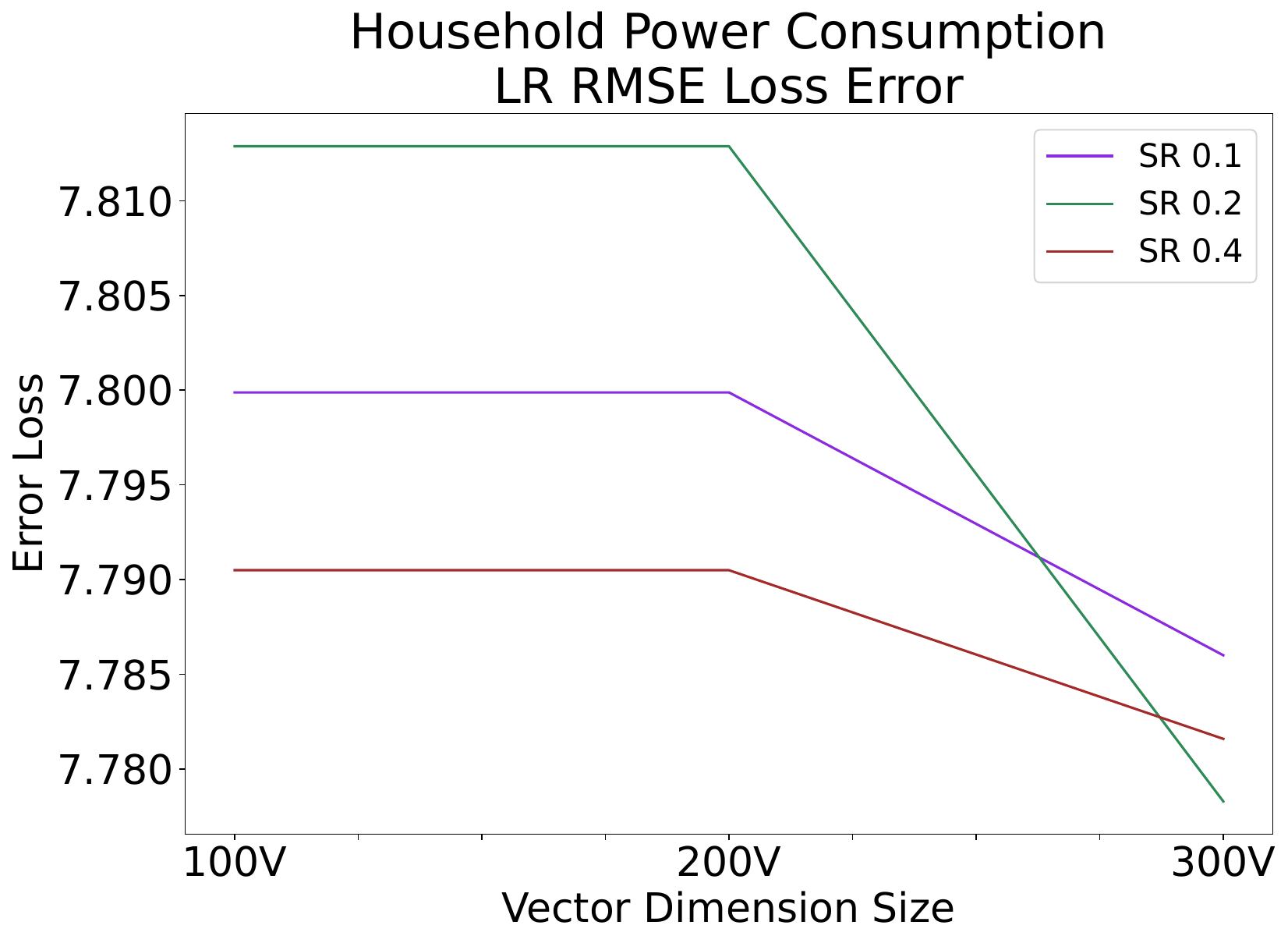}
         \caption{HPC dataset LR Operator RMSE error loss}
         \label{fig:Weather-SR-SVM-MAE}
     \end{subfigure}
     \caption{Sampling Ratio prediction results}
        \label{fig:SR-EVAL-RES}
\end{figure}

We conducted an experimental evaluation using the Sampling Ratio (SR) approach, similar to Apollo \cite{b7Apollo1}, but employed neural networks built from the vector embeddings of each dataset. The SR approach involves a unified random selection of $l\%$ datasets from the vector representation space, using this subset to construct a neural network for predicting operator outputs. We tested SR values of $0.1$, $0.2$, and $0.4$, as well as vector embedding dimensions of $100$, $200$, and $300$, across all datasets. 
Figure \ref{fig:SR-EVAL-RES} presents the sampling ratio results for the Adult dataset using MLP (sub-figure \ref{fig:Adult-SR-RMSE}) and for the Weather dataset using LR (sub-figure \ref{fig:Weather-SR-SVM-MAE}). In each sub-figure the y-axis represents the RMSE prediction error loss while the x-axis denotes the vector dimension

Both experiments demonstrate that as the vector embedding dimension increases, coupled with a larger sampling ratio (SR) value, there is a slight decrease in the prediction error loss. This improvement occurs because higher-dimensional vector embeddings provide a more accurate representation of the datasets in k-dimensions, with better dataset selection leading to enhanced prediction accuracy. Comparing the SR approach to our similarity search method for the HPC dataset, the SR approach was approximately $15\%$ less accurate in operator prediction across all vector embedding dimensions. A similar trend was observed in the Weather dataset. However, the Stock dataset exhibited a much larger discrepancy, with the SR approach performing about $70\%$ worse in prediction accuracy across all vector embedding dimensions. Likewise, in the Adult dataset, the SR approach delivered the poorest performance, with nearly $90\%$ lower prediction accuracy compared to the similarity search methods.

\begin{table}[!ht]
    \centering
    \vspace{-0.3in}
        \caption{Evaluation results of our framework exported analytic operator with lowest prediction error in comparison with Apollo}
        
    \label{tab:table-eval-res}
    \scalebox{0.7}{
    \setlength\doublerulesep{0.5pt}
    \begin{tabular}{|c|c|c|c|c|c|c|}
    \hline
         \makecell{Dataset\\Name} & Method & Operator & RMSE &  MAE & Speedup  & Amortized Speedup \\
         \hline\hline
         \multirow{7}{*}{\makecell{Household\\Power\\Consumption}}& \makecell{$300$V Cosine} & LR & $\mathbf{6.61}$ & $\mathbf{5.42}$ & $0.0017$ & $\mathbf{1.99}$ \\ \cline{2-7}
                  & \makecell{$300$V SR-$0.2$} & LR & $7.77$ & $6.66$ &  $0.0018$  & $1.42$\\ \cline{2-7} 
        & \makecell{Apollo-SR $0.1$} & LR & $2968.01$ &  $2352.55$ & $\mathbf{0.015}$ & $0.024$ \\ \cline{2-7}
         & \makecell{Apollo-SR $0.2$} & LR & $2811.49$ &  $2229.50$ & $0.015$ & $0.024$ \\ \cline{2-7}\cline{2-7}
         & \makecell{$300$V K-Means} & MLP Regr. & $\mathbf{6.70}$ & $\mathbf{3.38}$ &  $0.9249$  & $\mathbf{1.99}$\\ \cline{2-7}
         & \makecell{Apollo-SR $0.1$} & MLP Regr. & $3322.05$ &  $2606.99$ & $2.38$ & $1.74$ \\ \cline{2-7}
         & \makecell{Apollo-SR $0.2$} & MLP Regr. & $3850.01$ &  $2609.36$ & $\mathbf{2.38}$ & $1.74$\\ \cline{1-7} \cline{1-7} 
        \multirow{7}{*}{\makecell{Stock}} &  \makecell{$300$V Cosine} & LR & $306382.28$ & $125335.65$ & $0.00085$ & $\mathbf{1.91}$\\ \cline{2-7}
        & \makecell{$300$V SR-$0.4$} & LR & $21861625.91$ & $5674215.265$ &  $0.00087$  & $0.33$\\ \cline{2-7}
        & \makecell{Apollo-SR $0.1$} & LR & $\mathbf{153665.92}$ &  $\mathbf{118236.48}$ & $\mathbf{0.00093}$ & $0.00096$\\ \cline{2-7}
         & \makecell{Apollo-SR $0.2$} & LR & $166844.95$ &  $133306.68$ & $0.00093$ & $0.00096$\\ \cline{2-7}\cline{2-7}
         &  \makecell{$300$V Cosine} & MLP Regr. & $\mathbf{140236.47}$ & $\mathbf{123571.12}$ & $0.63$ & $\mathbf{1.91}$\\ \cline{2-7}
         & \makecell{Apollo-SR $0.1$} & MLP Regr. &  $175150.82$ &  $145123.09$ & $\mathbf{0.93}$ & $0.96$\\ \cline{2-7}
         & \makecell{Apollo-SR $0.2$} & MLP Regr. & $174390.81$ &  $146338.73$ & $0.93$ & $0.96$\\ \cline{1-7} \cline{1-7}
         \multirow{7}{*}{\makecell{Weather}}& \multirow{1}{*}{ \makecell{$300$V Cosine}} & \makecell{SVM SGD}& $\mathbf{14.13}$ & $\mathbf{7.63}$ & $1.06$ & $\mathbf{22.8}$ \\ \cline{2-7}
               & \makecell{Apollo-SR $0.1$} & SVM & $69.51$ &  $25.52$ & $\mathbf{2.10}$ &  $1.16$\\ \cline{2-7}
                        & \makecell{Apollo-SR $0.2$} & SVM & $68.70$ &  $22.81$ & $2.10$ & $1.16$\\ \cline{2-7} \cline{2-7}
       &  \multirow{1}{*}{ \makecell{$300$V Cosine}}& MLP & $\mathbf{14.29}$ & $\mathbf{4.03}$ & $1.03$  & $\mathbf{22.8}$\\ \cline{2-7}
        &  \multirow{1}{*}{ \makecell{$300$V SR-$0.4$}}& MLP & $15.95$ & $13.31$ & $1.02$  & $1.77$\\ \cline{2-7}
         & \makecell{Apollo-SR $0.1$} & MLP & $69.62$ &  $23.10$ & $\mathbf{1.34}$ & $1.14$ \\ \cline{2-7}
         & \makecell{Apollo-SR $0.2$} & MLP & $673.56$ &  $\mathbf{84.70}$ & $1.32$ & $1.14$\\ \cline{1-7} \cline{1-7}
         
         \multirow{7}{*}{\makecell{Adult}}& \multirow{1}{*}{ \makecell{$300$V Cosine}} & \makecell{SVM SGD}& $\mathbf{0.36}$ & $\mathbf{0.2}$ & $0.37$   & $\mathbf{2.78}$\\ \cline{2-7}
                  & \makecell{Apollo-SR $0.1$} & SVM & $68.32$ &  $22.95$ & $\mathbf{0.75}$ & $0.85$ \\ \cline{2-7}
                 & \makecell{Apollo-SR $0.2$} & SVM & $68.88$ &  $22.88$ & $0.74$ & $0.85$\\ \cline{2-7} \cline{2-7}

         &  \multirow{1}{*}{ \makecell{$300$V K-Means}}& MLP & $\mathbf{0.36}$ & $\mathbf{0.19}$ & $0.30$ & $2.78$ \\ \cline{2-7}
        & \makecell{$300$V SR-$0.2$} & MLP & $6.01$ & $6.00$ &  $0.54$  & $\mathbf{3.54}$\\ \cline{2-7}
         & \makecell{Apollo-SR $0.1$} & MLP & $71.11$ &  $26.51$ & $\mathbf{1.07}$ & $1.31$\\ \cline{2-7}
         & \makecell{Apollo-SR $0.2$} & MLP & $70.16$ &  $25.74$ & $1.05$ & $1.31$\\ \cline{1-7}
         
    \end{tabular}
    }
    \vspace{-0.3in}
\end{table}

For the HPC and Weather datasets, the SR approach was approximately $15\%$ less accurate in operator prediction compared to all similarity search methods, even as vector embedding dimensions increased. In contrast, the Stock dataset exhibited a significantly larger discrepancy, with the SR approach performing about $70\%$ worse in prediction accuracy across all vector embedding dimensions. Similarly, in the Adult dataset, the SR approach recorded the poorest performance, delivering nearly $90\%$ worse prediction accuracy compared to the similarity search methods.

Table \ref{tab:table-eval-res} compares model operators, loss functions, and speedup metrics for our framework and Apollo at SR values of $0.1$ and $0.2$. Methods $100$V, $200$V, and $300$V denote vector embedding dimensions. The lowest prediction errors align with our pipeline's similarity search method.
Apollo outperforms our framework on the Stock dataset for the LR analytic operator with the smallest amount of SR. However, our framework excels with the MLP regression operator, improving RMSE and MAE by $20\%$ and $17\%$, respectively. The LR operator's performance gap on the Stock dataset is minor.
For other datasets, our framework consistently surpasses Apollo across different SR values. This demonstrates the effectiveness of our similarity search approach, which enhances data quality and reduces $\Phi$ prediction errors by identifying relevant datasets $D_r$ from the data lake directory $D$. The Adult dataset also highlights our framework's advantage with increasing feature dimensions.
Although Apollo achieves better raw speedup due to the higher complexity of our framework's vectorization step, our framework outperforms it in amortized speedup. By excluding the reusable vectorization process, it achieves speed gains of $10\%$ to $60\%$ for most operators.
The SR approach, leveraging vector embedding representations, enhances operator prediction compared to Apollo and achieves greater amortized speedup. However, the similarity search method outperforms both Apollo and the SR approach in prediction accuracy and amortized speedup, establishing its clear superiority across most datasets and operator scenarios.

\vspace{-0.2in}
\subsection{NumTabData2Vec Evaluation Results}

\begin{table}[!ht]
    \centering
    \vspace{-0.5in}
    \caption{Similarity between vectors of different datasets scenarios}
    \scalebox{0.8}{
    \label{tab:vec-rep-sim}
    \setlength\doublerulesep{0.5pt}
    \begin{tabular}{||c|c||}
    \hline
    \makecell{\textit{NumTabData2Vec}\\Vector Size} & Similarity \\
    \hline\hline
     \makecell{$100$} & $0.54$\\
     \hline
      \makecell{$200$}   & $0.18$\\
      \hline
       \makecell{$300$}  & $0.16$\\ \hline
    \end{tabular}}
    \vspace{-0.5in}
\end{table}

\begin{multicols}{2}
    \begin{figure*}[!ht]
    \vspace{-0.2in}
    \centering
    \begin{minipage}{0.48\textwidth} 
        \centering
        \includegraphics[width=\textwidth]{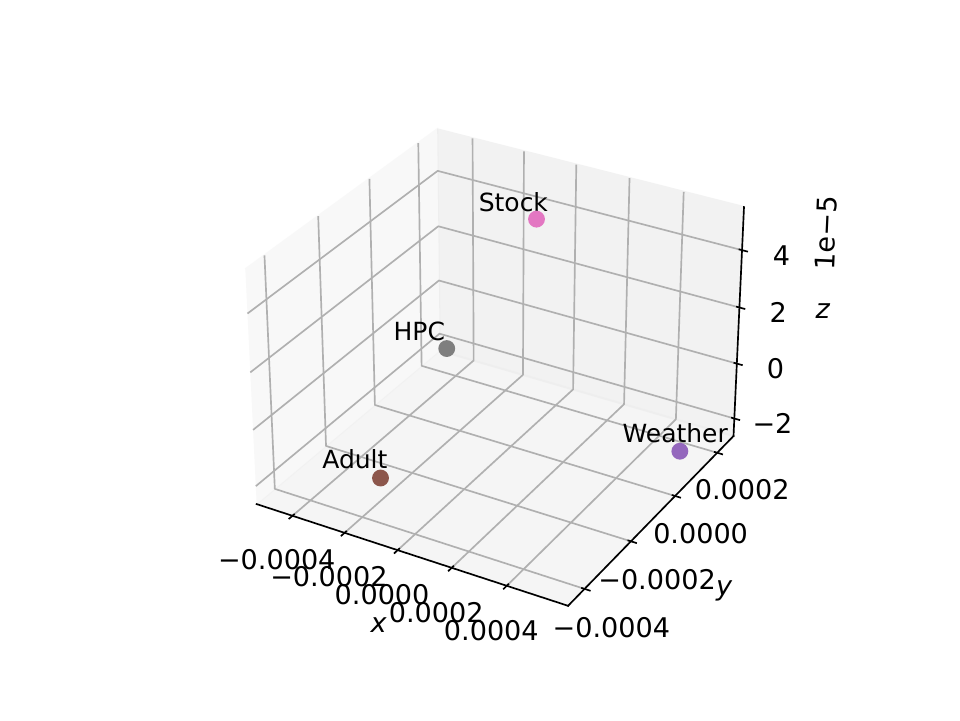}
        \caption{Vector representation for each dataset from NumTabData2Vec}
        \label{fig:eval-data-repr}
         \vspace{-0.3in}
    \end{minipage}
    \hfill
    \begin{minipage}{0.48\textwidth} 
        \centering
        \vspace{0.2in}
        \includegraphics[width=\textwidth]{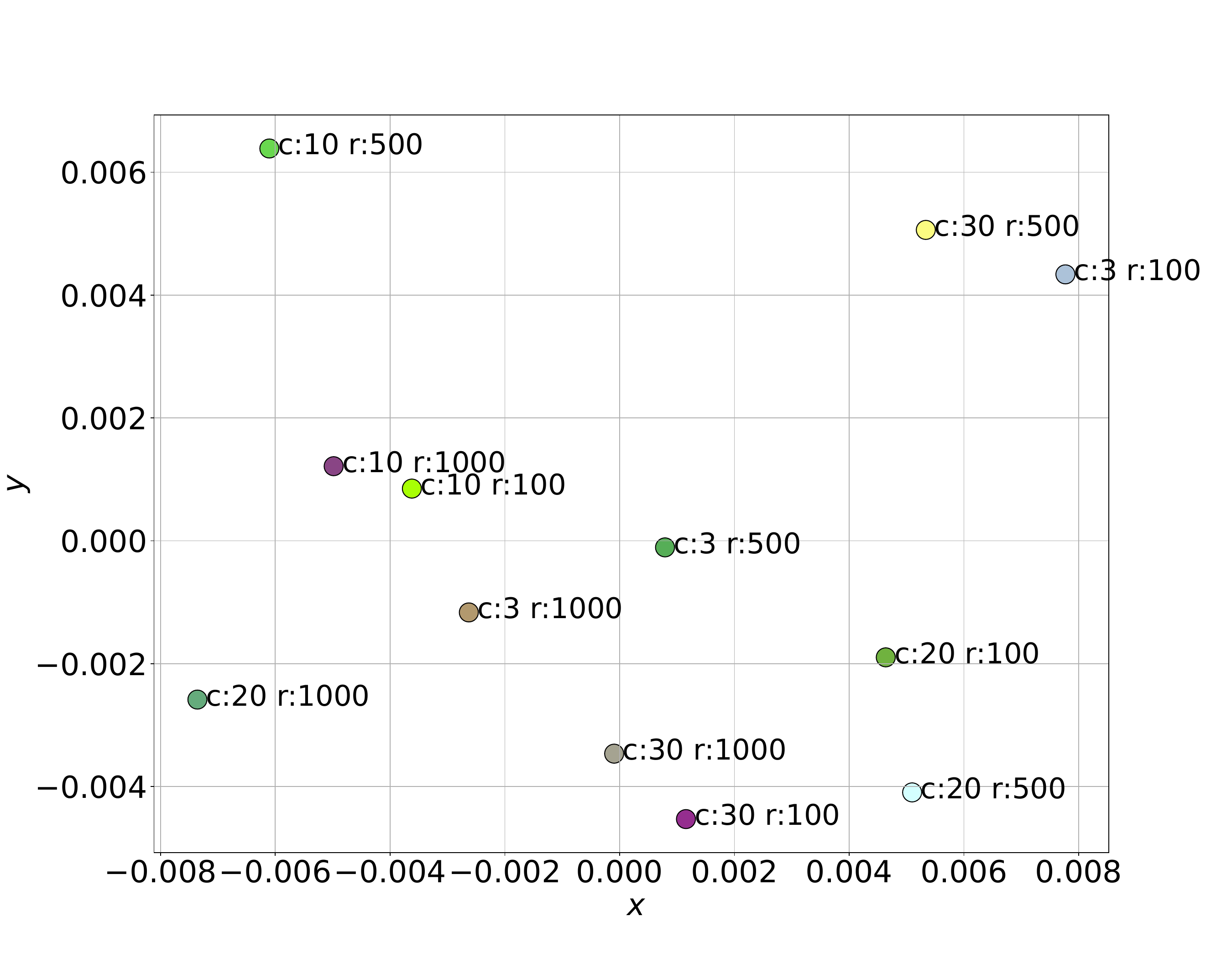}
        \vspace{-0.4in}
        \caption{Synthetic data vector embedding representation}
        \vspace{-0.3in}
        \label{fig:eval-sd-data-repr}
    \end{minipage}
\end{figure*}
\end{multicols}

Our proposed model, \textit{NumTabData2Vec},was evaluated for its ability to distinguish dataset scenarios based on qualitative differences. For each scenario, $n$ random datasets were reduced to 3D vector embeddings using PCA, as shown in Figure \ref{fig:eval-data-repr}, which demonstrates \textit{NumTabData2Vec}'s ability to distinguish datasets with minimal overlap across contexts. Unlike prior methods \cite{b8Word2Vec, b9Graph2Vec}, which focus on text or graphs, \textit{NumTabData2Vec} applies to entire datasets. Table \ref{tab:vec-rep-sim} further highlights the average cosine similarity between dataset embeddings, showing greater dissimilarity as vector dimensions increase. However, results suggest that dimensions between $100$ and $300$ are sufficient for accurate distinction, avoiding the need for larger vector sizes.

\begin{table}[!htb]
    \vspace{-0.4in}
    \centering
    \caption{NumTabData2Vec execution time for different dataset dimensions and different vector sizes }

    \begin{adjustbox}{width=\columnwidth,center}
    \label{tab:vec-exec-time}
    \setlength\doublerulesep{0.5pt}
    \begin{tabular}{||c|c|c|c|c||}
    \hline
     \makecell{\# of columns} & \makecell{\# of rows} & \makecell{$50$ Vectors\\Execution time} & \makecell{$100$ Vectors\\Execution time} & \makecell{$200$ Vectors\\Execution time} \\
    \hline\hline
     $3$ & $100$ & $0.0004$ sec & $0.00042$ sec & $0.00051$ sec\\ \hline
     $3$ & $500$ & $0.0004$ sec & $0.00041$ sec & $0.00049$ sec\\ \hline
     $3$ & $1000$ & $0.0004$ sec & $0.00041$ sec & $0.00049$ sec\\ \hline
     $3$ & $1500$ & $0.0004$ sec & $0.00041$ sec & $0.00055$ sec\\ \hline
     $3$ & $1800$ & $0.0004$ sec & $0.00041$ sec & $0.00055$ sec\\ \hline
     \hline
     $10$ & $100$ & $0.0004$ sec & $0.0004$ sec & $0.00057$ sec\\ \hline
     $10$ & $500$ & $0.00039$ sec & $0.0004$ sec & $0.00051$ sec\\ \hline
     $10$ & $1000$ & $0.00041$ sec & $0.00042$ sec & $0.00052$ sec\\ \hline
     $10$ & $1500$ & $0.00041$ sec & $0.00042$ sec & $0.00055$ sec\\ \hline
     $10$ & $1800$ & $0.00041$ sec & $0.00042$ sec & $0.00052$ sec\\ \hline
     \hline
     $20$ & $100$ & $0.0004$ sec & $0.00042$ sec & $0.0005$ sec\\ \hline
     $20$ & $500$ & $0.0004$ sec & $0.00042$ sec & $0.0005$ sec\\ \hline
     $20$ & $1000$ & $0.00042$ sec & $0.00043$ sec & $0.00052$ sec\\ \hline
     $20$ & $1500$ & $0.00043$ sec & $0.00044$ sec & $0.00054$ sec\\ \hline
     $20$ & $1800$ & $0.00044$ sec & $0.00044$ sec & $0.00054$ sec\\ \hline    
     \hline\hline
    \end{tabular}
    \end{adjustbox}
\end{table}

\begin{multicols}{2}
    \begin{figure*}[!ht]
    \vspace{-0.2in}
    \centering
    \begin{minipage}{0.47\textwidth} 
        \centering
        \includegraphics[width=\textwidth]{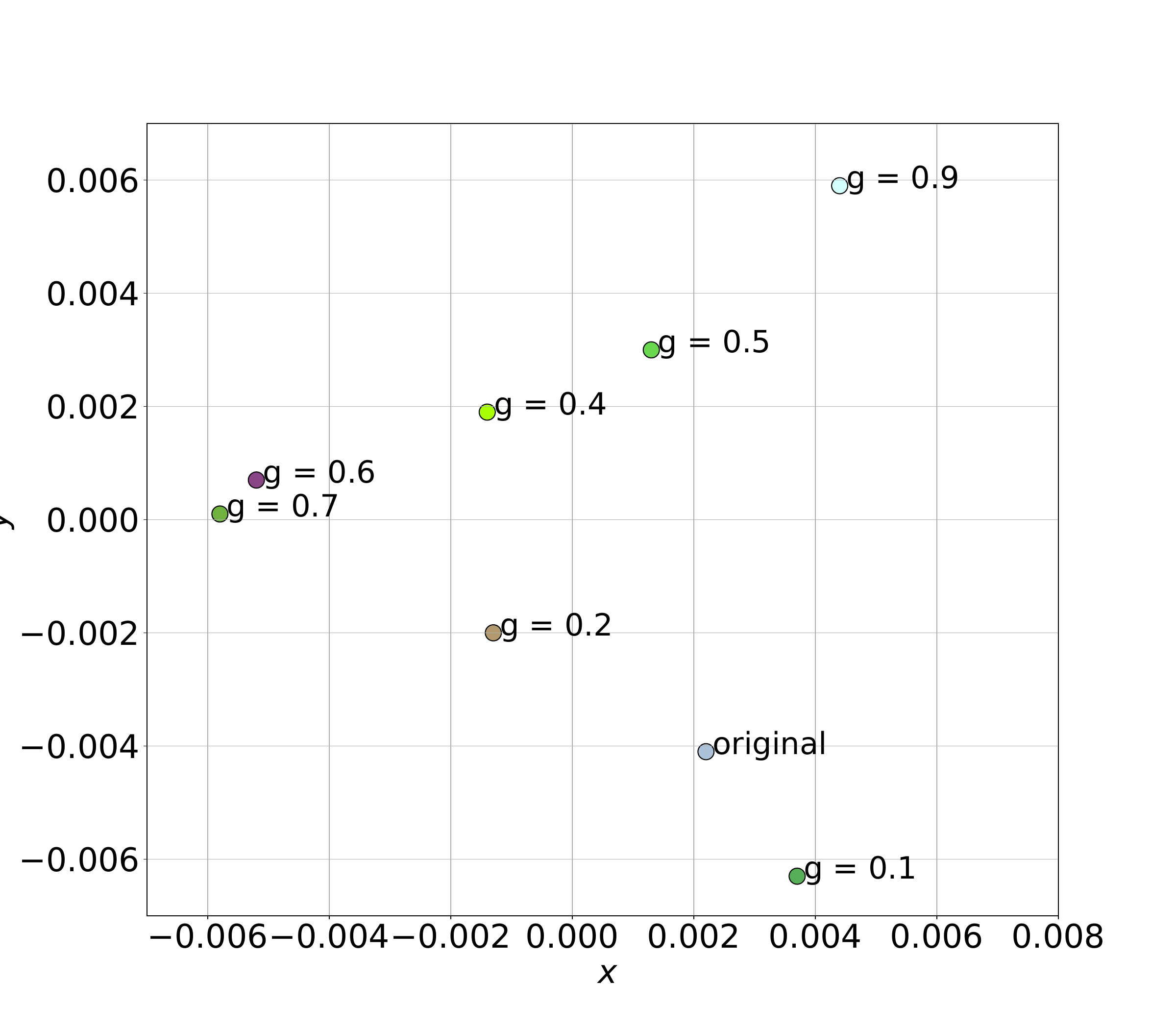}
        \vspace{-0.35in}
        \caption{HPC Dataset vector embedding representation with addition of Noise}
        \label{fig:eval-nd-data-repr}
        \vspace{-0.35in}
    \end{minipage}
    \hfill
    \begin{minipage}{0.47\textwidth} 
        \centering
        \includegraphics[width=\textwidth]{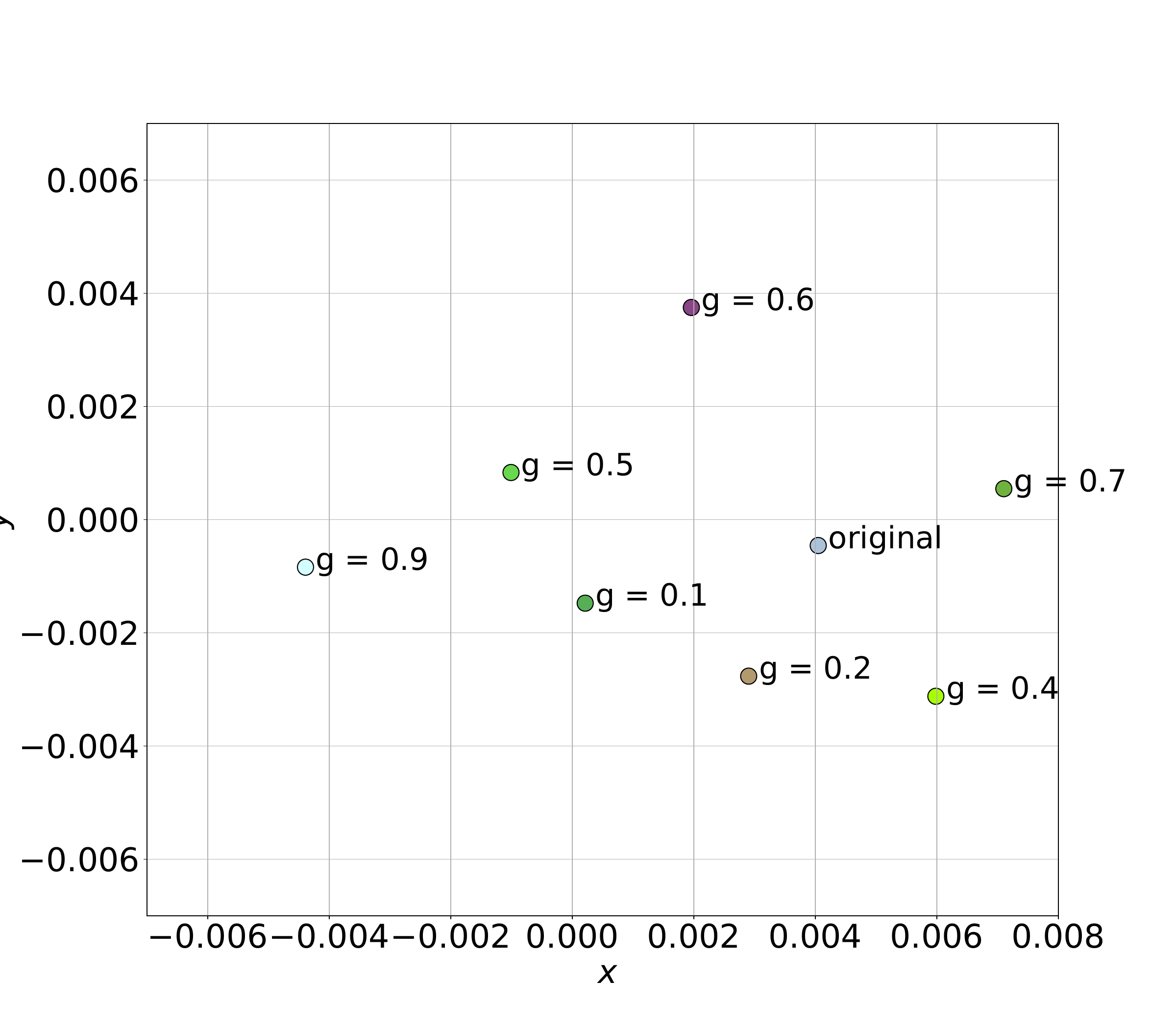}
         \vspace{-0.35in}
        \caption{HPC Dataset vector embedding representation with addition of Noise in the first column}
        \label{fig:eval-nd-data-repr-1col}
        \vspace{-0.35in}
    \end{minipage}
\end{figure*}
\end{multicols}

To evaluate \textit{NumTabData2Vec}'s ability to distinguish datasets with varying row and column counts, we generated synthetic numerical tabular datasets of different dimensions and vectorized them. Figure \ref{fig:eval-sd-data-repr} shows datasets with columns ranging from three to thirty and rows from ten to one thousand, projected from a $200$-dimensional space to 2D using PCA. Each bullet caption c and r corresponds to the columns and rows of the dataset, respectively. Datasets with the same number of columns cluster closely in the representation space, and a similar pattern is observed for datasets with the same number of rows. These results indicate that our method effectively distinguishes datasets based on size during vectorization.

To evaluate \textit{NumTabData2Vec}'s ability to distinguish datasets by distribution and order, Gaussian noise was added to $l\%$ of tuples in an HPC dataset. Figure \ref{fig:eval-nd-data-repr} shows the original and noise-modified datasets projected to 2D using PCA, with greater noise causing larger shifts in the representation space. This demonstrates the model's effectiveness in capturing distribution differences and distinguishing datasets based on ordering.

To assess fine-grained distinctions, we repeated the experiment by adding Gaussian noise exclusively to the first column of the dataset. Figure \ref{fig:eval-nd-data-repr-1col} shows the 2D vector space, where g in the bullet caption indicates the noise level. As noise increases, the representation shifts further from the original dataset, though it remains closer than in Figure \ref{fig:eval-nd-data-repr}, with points more tightly grouped in 2D space.

To evaluate how dataset dimensions affect the execution time of \textit{NumTabData2Vec}, we created synthetic datasets with varying numbers of rows ($100$, $500$, $1000$, $1500$, and $1800$) and columns ($3$, $10$, and $20$). These datasets were vectorized into different dimensions, and the execution times were recorded. Table \ref{tab:vec-exec-time} shows that increasing the k-dimension requires approximately $20\%$ more time to generate the vector embeddings. This is expected, as a higher k-dimension involves more hyperparameters, which naturally increases computation time.

Interestingly, varying the number of columns did not significantly impact execution time. However, increasing the number of rows resulted in approximately $5\%$ additional execution time. This is because larger datasets require the extraction of more features, which has a modest impact on the model's execution time.

\vspace{-0.2in}
\section{Limitations}
\vspace{-0.15in}
There exist a number of limitations in our work as we described it. In this section we briefly highlight them.
Firstly, our input datasets comprise records of specific size and type (numerical). This currently excludes data with textual and categorical attributes, or tables with varying number of features inside a set of datasets.
Secondly, we currently consider single-input and single-output operator modelling. 
Finally, our proposed NumTabData2Vec model for data vectorization has a performance limitation, as it cannot deal with datasets bigger than about $3000$ tuples. This is mostly a hardware limitation of off-the-shelf GPUs (with at most 24GB of memory available for a budget GPU).

\vspace{-0.2in}
\section{Conclusion}
\vspace{-0.15in}
In this paper, we presented a novel framework for the modelling of an analytic operator (such as a ML algorithm) when a large number of input data is available and thus no brute-force execution can be performed. We propose a deep learning model, \textit{NumTabData2Vec}, which transforms a dataset to a lower $k$-dimensional representation space $z$. Our framework produces vector embeddings for the input datasets using \textit{NumTabData2Vec} and performs a similarity search to identify the most relevant subset of datasets for any unseen input. By modelling the analytic operator based on this selected subset, we are able to accurately predict its output on any given input dataset. In practice, we demonstrated that our framework can accurately model various common algorithms and compared favourably against a similar recent framework \cite{b7Apollo1}, in both accuracy and speedup. Furthermore, we showed that \textit{NumTabData2Vec} can create different vector representations for datasets from different scenarios. We also demonstrated that \textit{NumTabData2Vec} can effectively detect when noise is introduced into a dataset.

\vspace{-0.15in}
\begin{credits}
\subsubsection{\ackname} This work is partially supported by the project RELAX-DN, funded by the European Union under Horizon Europe 2021-2027 Framework Programme Grant Agreement number 101072456.
\vspace{-0.2in}

%

\end{credits}
%
%
%
\bibliographystyle{splncs04}
\bibliography{refs}%




\end{document}